\documentclass[preprint,1p,10pt,authoryear]{elsarticle}

\usepackage{graphicx}
\usepackage{amsmath,amsfonts,amssymb}
\usepackage[ruled]{algorithm}
\usepackage{algpseudocode} 
\usepackage{hyperref}
\usepackage{xcolor}
\usepackage{wrapfig}

\newcommand{\Real}{\mathbb R}                                   
\newcommand{\set}[1]{\left\{#1\right\}}                         
\newcommand{\cset}[2]{\left\{#1\,\left\vert\,#2\right.\right\}} 
\newcommand{\ceil}[1]{\left\lceil #1 \right\rceil}              
\newcommand{\floor}[1]{\left\lfloor #1 \right\rfloor}           
\newcommand{\norm}[1]{\left\Vert#1\right\Vert}                  

\newcommand{\transp}[1]{#1^\top}                              
\newcommand{\concat}[1]{[#1]}                                   
\newcommand{\concatt}[1]{\transp{\concat{#1}}}                  

\newcommand{\aux}[1]{\tilde{#1}}                                    
\def\argmax{\operatornamewithlimits{arg\,max}}                  
\def\argmin{\operatornamewithlimits{arg\,min}}                  
\newcommand{\eps}{\varepsilon}                                  

\newcommand{\x}{x}

\renewcommand{\u}{u}                                            

\renewcommand{\k}{k}                                            

\renewcommand\d{d}
\renewcommand\b{b}

\newcommand{\opt}[1]{#1^{*}}                                    
\newcommand{\p}{\theta}                                         
\renewcommand{\b}{b}                                            

\newcommand{\algorithmicinput}{\textbf{Input:}}

\def\Input{\item[\algorithmicinput]}

\newcommand{\secref}[1]{Section~\ref{#1}}
\newcommand{\figref}[1]{Figure~\ref{#1}}
\newcommand{\tabref}[1]{Table~\ref{#1}}
\newcommand{\algoref}[1]{Algorithm~\ref{#1}}

\newcounter{example}
\newtheorem{exampleaux}[example]{Example}
\newenvironment{example}[1][]
    {\vspace*{.2em}\begin{exampleaux}\textit{#1}
    \normalfont}
    {\hfill $\square$\end{exampleaux}}

\newcommand\rev[1]{#1}
\newcommand\rerev[1]{{\color{blue} #1}}

\journal{Control Engineering Practice}

\begin{document}

\begin{frontmatter}

\title{Learning control for transmission and navigation with a mobile robot under unknown communication rates}


\author[utcluj]{Lucian Bu\c{s}oniu}
\author[cran]{Vineeth S.\ Varma}
\author[cran]{J\'er\^ome Loh\'eac}
\author[utcluj]{Alexandru Codrean}
\author[upt]{Octavian \c{S}tefan}
\author[cran]{Irinel-Constantin Mor$\breve{\text{a}}$rescu}
\author[cran,supelec]{Samson Lasaulce}

\address[utcluj]{Technical University of Cluj-Napoca, Memorandumului 28, 400114 Cluj-Napoca, Romania\\ lucian@busoniu.net (corresponding author), alexandru.codrean@aut.utcluj.ro}
\address[cran]{Universit\'e de Lorraine, CNRS, CRAN, F-54000 Nancy, France. \{vineeth.varma, jerome.loheac, constantin.morarescu\}@univ-lorraine.fr}
\address[upt]{Automation and Applied Informatics, Politehnica University of Timisoara, Romania\\octavian.stefan@aut.upt.ro}
\address[supelec]{CentraleSupelec - Univ. Paris Sud, Gif-sur-Yvette, France\\samson.lasaulce@centralesupelec.fr}

\begin{abstract}
In tasks such as surveying or monitoring remote regions, an autonomous robot must move while transmitting data over a wireless network with unknown, position-dependent transmission rates. For such a robot, this paper considers the problem of transmitting a data buffer in minimum time, while possibly also navigating towards a goal position. Two approaches are proposed, each consisting of a machine-learning component that estimates the rate function from samples; and of an optimal-control component that moves the robot given the current rate function estimate. \rev{Simple obstacle avoidance is performed for the case without a goal position. In extensive simulations, these methods achieve competitive performance compared to known-rate and  unknown-rate baselines.} A real indoor experiment is provided in which a Parrot AR.Drone 2 successfully learns to transmit the buffer.
\end{abstract}


\end{frontmatter}

\section{Introduction}
\label{sec:intro}

This paper considers problems in which a mobile robot must transmit a data file (or equivalently, empty a data buffer) over a wireless network, with transmission rates that depend on the robot position and may be affected by noise. The objective is to move in such a way that the buffer is emptied in minimum time. Two versions will be considered: one in which there is no desired end position for the robot, which is called the transmission problem and abbreviated as PT; and a second version in which the trajectory must end at a given goal position, called navigation-and-transmission problem (PN). Such problems appear e.g.\ when a UAV autonomously collects survey data (video, photographic, etc.) which it must then deliver over an ad-hoc network, possibly while navigating to the next mission waypoint.

The key challenge is that the rate function is usually unknown to the robot, e.g. because depends on unknown propagation environment effects like path loss, shadowing, and fast fading. For both PT and PN, algorithms are proposed that (i) learn approximations of the rate function from values sampled so far along the trajectory and (ii) at each step, apply optimal control with the current approximation to choose robot actions. In PT, (i) is done with supervised, local linear regression \citep{MooreAtkeson:93} and (ii) with a local version of dynamic programming \citep{Bertsekas:12}, for arbitrarily-shaped rate functions that are assumed deterministic for the design. \rev{The PT method includes a simple obstacle avoidance procedure.} In PN, component (i) uses active learning \citep{Settles:09} and (ii) is a time-optimal control design from \citet{Loheacetal:19}. In the latter case, the rate function is learned via the signal-to-noise ratio, which is affected by random fluctuations and is taken to have a radially-symmetric expression with unknown parameters.

A common idea in both the PT and PN approaches is to focus the learning method on the unknown element of the problem: the rate function, while exploiting the known motion dynamics of the robot in order to achieve the required fast, single-trajectory learning. Both approaches are validated in extensive simulations for a robot with nonlinear, unicycle-like motion dynamics and where the rate functions are given by single or multiple antennas with a path-loss form. Moreover, a real indoor experiment is provided that illustrates PT with a single router and a Parrot AR.Drone 2.

The problem of optimizing sensing locations to reduce the uncertainty of an unknown map is well-known in machine learning, see e.g.\ \citet{Krauseetal:08} for some fundamental results. This problem is often solved dynamically, to obtain so-called informative path planning, see e.g.\ \citet{Meeraetal:19} and \citet{Viserasetal:16}; \citet{Popovicetal:18} provide a good overview of this field. Closer to the present work, \citet{FinkKumar:10, Penumarthietal:17} learn a radio map with multiple mobile robots. Compared to these works, the methods used in this paper to learn the map are much simpler --- basic regression methods, whereas Gaussian processes \citep{RasmussenWilliams:06} are often used in the references above. However, the novelty here is that the robot has \emph{control and communication objectives}, whereas in most of the the works above the objective of the robot is just to learn the map (keeping in mind that learning must typically be done from a small number of samples, which is related but not identical to the time-optimality objectives in the present paper).

On the other hand, a related thread of work in engineering, which does consider joint control and communication objectives, is motion planning under connectivity, communication rate, or quality-of-service constraints \citep{Finketal:13, GhaffarkhahMostofi:11, RookerBirk:07, Chatzipanagiotisetal:12}. Closer to the present work, when the model of the wireless communication rate is known, some recent works \citep{Ooietal:09, Liceaetal:16, Gangulaetal:17, Loheacetal:19} have optimized the trajectory of the robot while ensuring that a buffer is transmitted along the way. The key novelty here with respect to these works is that the rate function is \emph{not known in advance} by the robot. Instead, it must be learned from samples observed while the robot travels in the environment. \rev{In particular, the approach to solve PN incorporates the known-rate method of \citet{Loheacetal:19} by applying it iteratively, at each step, for the current learned estimate of the rate function.}

Some recent works \citep{Finketal:12, Liceaetal:17, Rizzoetal:13, Rizzoetal:19} explore control of robots with rate functions that are uncertain but still have a known model. \rev{For instance, \citet{Rizzoetal:13, Rizzoetal:19} carefully develop rate models for tunnels, and robots then adapt to the parameters of these models.} In \citet{YanMostofi:12}, the rate function is initially unknown, but the trajectory of the robot is fixed and only the velocity is optimized.

\rev{Compared to the preliminary conference paper \citep{acc19}, fully novel contributions here are solving PN with unknown rates and the real-system results. \citet{acc19} focused only on a simpler version of PT, with first-order robot dynamics, synthetic rate functions, and without obstacles. For the PT version studied here, the robot dynamics are extended to be of arbitrary order, the method is illustrated with realistic rate functions, and -- importantly -- basic obstacle avoidance functionality is introduced.}


Next, \secref{sec:problem} gives the problem definition, following which the paper treats separately PT (\secref{sec:transmit} for the algorithm and \secref{sec:transmit:results} for the results) and PN (Sections \ref{sec:nav} and \ref{sec:navresults}, respectively). The real-system illustration is provided in \secref{sec:realtime}, and \secref{sec:conclusions} concludes the paper.


\section{Problem definition}\label{sec:problem}

Consider a mobile robot with position $p \in P \subseteq \Real^2$, additional motion-related states $y \in Y \subseteq \Real^{n_y}$, $n_y \geq 0$, and inputs $u \in U \subseteq \Real^{n_u}$, $n_u \geq 1$. The extra states $y$ may contain e.g.\ velocities, headings, or other variables needed to model the robot's motion. Dimension $n_y$ may be zero, in which case the only state signals are the positions and the robot has first-order dynamics. In this case, variable $y$ can be omitted from the formalism below (and, by convention, $\Real^0$ is a singleton). A discrete-time setting is considered with $k$ denoting the time step, so the robot has motion dynamics $g: P\times Y \times U \to P \times Y$:
\begin{equation}\label{eq:motiondynamics}
\concatt{\transp p_{k+1}, \transp y_{k+1}} = g(\concatt{\transp p_k, \transp y_k}, u_k)
\end{equation}
To apply e.g.\ the dynamic programming procedure of \secref{sec:transmit}, dynamics \eqref{eq:motiondynamics} should be Lipschitz-continuous.

The robot also carries a \emph{data buffer} of size $b \in \Real_+$ that it must empty over a wireless network with a transmission rate that varies with the position and may be affected by random fluctuations. The rate at step $k$ is $r_k \sim \mathcal{R}(p_k)$, where $\mathcal{R}$ is a position-dependent density function. Here, $r_k \in \Real$, $r_k \geq 0$, and notation `$\sim$' means `sampled from'. Therefore, the buffer size evolves like:
\begin{equation}\label{eq:bufferdyn}
b_{k+1} = \max \set{0, b_k - T_{s} r_k}
\end{equation}
where $T_s$ is the sampling period. Note that to obtain a discrete-time dynamics consistent with \eqref{eq:motiondynamics}, $r_k$ is taken constant along the  sampling period. This $r_k$ may be interpreted as a continuous, average rate, leading also to a continuous buffer size $b$, which is convenient in the algorithms.

Denote the overall state by $x := \concatt{\transp p, \transp y, b} \in X$, $X := P \times Y \times \Real_+$, containing the position, the extra motion states, and the buffer size. Thus, the overall dynamics are:
\begin{equation}\label{eq:f}
x_{k+1} = f(x_k, u_k, r_k) := \begin{bmatrix}g(\concatt{\transp p_k, \transp y_k}, u_k) \\ \max \set{0, b_k - T_s r_k}\end{bmatrix}
\end{equation}

Two related problems shall be considered.

\begin{itemize}
\item[PT] \label{pt}
(Transmission Problem) Given an initial state $x_0 = \concatt{\transp p_0, \transp y_0, b_0}$, deliver the buffer in minimum time, \rev{possibly in the presence of obstacles}.
\item[PN] \label{pn}
(Navigation and transmission Problem) Given $x_0 = \concatt{\transp p_0, \transp y_0, b_0}$, drive the robot to a goal position $\opt p$ in minimum time, in such a way that buffer delivery is complete when $\opt p$ has been reached.
\end{itemize}

To design the algorithm for PT, the transmission rate will be required to be deterministic. \rev{Define thus function $R : P \to [0, \infty)$, so that $r_k = R(p_k)$.} Contrast this to the general case above, in which $r_k \sim \mathcal{R}(p_k)$ with $\mathcal{R}$ the density function (keeping in mind that in the upcoming numerical experiments, the algorithm performs well even for random rates). The shape of $R$ is arbitrary, so e.g.\ multiple transmitters/receivers may exist, informally called ``antennas'' throughout the paper. \rev{Regarding the obstacles, it will assumed that there is a finite number of them and that the initial state is initialized sufficiently far away that the robot can still avoid the obstacle. Recall that in the version of PT solved in \citet{acc19}, there were no obstacles, and the dynamics had to be first-order, without any signal $y$.}


In PN, it will be assumed that the dynamics are first-order ($n_y=0$). A second, key assumption is that there is a single antenna around which the rate function decreases radially, see below for some specific formulas. This is a standard setting in the communication literature. The rationale is that given these restrictions, and for known and deterministic rate functions, \citet{Loheacetal:19} provide a design procedure for the optimal control in a continuous-time version of PN, which will later be exploited in the algorithm. \rev{When there are multiple antennas, the design of \citet{Loheacetal:19} does not work anymore. An extension may nevertheless be imagined in which that design may e.g.\ be heuristically applied to the closest (or strongest) antenna. This extension is left for future work.}

The signal-to-noise ratio (SNR) of the transmission channel equals (in expected value):
\begin{equation}\label{eq:SNR}
S(p_k) = \frac{K}{(\norm{p_k - p_{\mathrm{ant}}}+h)^\gamma}
\end{equation}
where $K$ is a constant that includes both the transmission power (assumed constant throughout the paper) and a normalization factor, and $p_{\mathrm{ant}}$ is the antenna position. Both $h$ and $\gamma$ influence the speed at which the SNR decreases; $\gamma$ is the path loss exponent. Parameters $K, h, \gamma$ are positive. Notation $\norm{\cdot}$ indicates the Euclidean norm.

The expected value \eqref{eq:SNR} is multiplied with a random number $z$ drawn from a Rice density, leading to the so-called Rician fading:
\begin{equation}\label{eq:rice}
z_k = \frac{1}{E_z}\norm{\begin{bmatrix}z' + v \\ z''\end{bmatrix}}
\end{equation}
where $z'$ and $z''$ are two independent, zero-mean, unit-variance, normally distributed random numbers; $v$ is a parameter that dictates the variance of $z$ (when $v$ is smaller, $z$ varies more widely); and $E_z$ is a normalization coefficient that ensures the expected value of $z$ is $1$.

Finally, the rate at step $k$ is given by:
\begin{equation}\label{eq:Rpath}
r_k = R_0 \log_2 [1 + z_k S(p_k)]
\end{equation}
with $R_0$ an application-dependent constant. This expression implicitly gives the rate density function $\mathcal{R}(p)$ from above. \rev{Note also that \eqref{eq:Rpath} is an ideal form, which only upper-bounds the rates in practice; so in fact, here an additional simplification is made, by requiring that the rates are (at least close to) the ideal ones.}

\rev{The specific forms \eqref{eq:SNR} and \eqref{eq:Rpath} are taken largely for convenience. The algorithm can be easily extended to any parameterized form of the SNR and to any relationship between the SNR and the rate, with the key overall requirement that rates must radially decrease around the antenna to be able to apply the control design of \citet{Loheacetal:19}.}

\medskip
In both PT and PN, the contribution is motivated by the fact that in practice the rate function $R$ (given in PN via the SNR $S$) is unknown to the robot, because it depends on propagation environment effects. The main objective in the sequel is, therefore, to derive efficient algorithms for unknown rate or SNR functions. The robot only has access to realizations at particular positions. For instance, rates may be measured via ACK/NACK feedback mechanisms. Algorithms to measure SNR are standard even in consumer devices. The robot can therefore accurately sample the rate or SNR once it reaches position $p_k$ and can use this information to make decisions at step $k$.


The problem of learning control solutions when the dynamics are unknown is -- among other fields -- the focus of the reinforcement learning (RL) \citep{SuttonBarto:18}. However, the problem considered here is quite different from standard RL: while the typical paradigm is that RL is applied across many trajectories, seeing the same or similar states over and over again, here one cannot afford to wait several trajectories for good performance: indeed, the robot is only given a single trajectory to transmit its data. Performance is only important during this trajectory, and for only those states encountered along it, most of which will be seen only once. The second difference is that here significant information is available about the model: with the exception of the rates, everything is known in $f$ in \eqref{eq:f}. The key idea is to exploit the second difference in order to address the challenge stemming from the first; that is, to learn rates or SNRs directly and use their estimate in $f$ to achieve fast, single-trajectory learning. 

In particular, for PT, a procedure will be designed that combines supervised learning to estimate $R$, with dynamic programming to find an approximation of \eqref{eq:hstar}. For PN, active learning of the SNR function will be combined with optimal control redesign at each step (for the current estimate of the SNR) using the procedure of \citet{Loheacetal:19}.

\rev{
So, to sum up, in this paper two problems are solved: PT (in \secref{sec:transmit}) and PN (in \secref{sec:nav}). While the main focus is the unknown-rate version of these two problems, a model-based, known-rate procedure is still required for each. Thus, for PT a known-rate dynamic programming procedure is designed first, in \secref{sub:transmit:dp}, followed by the unknown-rate algorithm in \secref{sub:transmit:learn}. Similarly, for PN, the known-rate procedure of \citet{Loheacetal:19} is presented in \secref{sub:nav:mb}, followed by the novel unknown-rate approach in \secref{sub:nav:learn}.

To get insight on the relationship between the two problems, note first that PT is, in general, simpler than PN, since the latter has an additional navigation objective. Nevertheless, due to the different assumptions made, the actual relationship is more intricate. In PT, the approach handles fully unknown, arbitrarily-shaped but deterministic rate functions $R$. In PN, the SNR function is unknown and affected by random fluctuations; it must however radially decrease around an antenna position. Thus, a more accurate statement of the relationship is that PN extends PT to navigation objectives and random fluctuations, but at the cost of imposing radial rates around a single antenna.

An easy to visualize roadmap is shown in Table \ref{tab:roadmap}.
}

\begin{table}[!htb]
  \centering
  \rev{\begin{tabular}{|p{2cm}|p{5cm}|p{5cm}|}
    \hline
      & PT & PN \\\hline
    Rate type & Deterministic, arbitrary shape & Random, radial around one antenna \\\hline
    Known-rate method   & DP, Sec.~\ref{sub:transmit:dp} & optimal control from \citet{Loheacetal:19}, Sec.~\ref{sub:nav:mb}\\\hline
    Unknown-rate method & DP + supervised learning of $R$, Sec.~\ref{sub:transmit:learn} & as above + active learning of SNR, Sec.~\ref{sub:nav:learn}\\
    \hline
  \end{tabular}}
  \caption{Roadmap of problems and techniques described in the paper}\label{tab:roadmap}
\end{table}


\section{Solution for the transmission problem}\label{sec:transmit}

First, PT is reformulated as a deterministic optimal control problem. \rev{Define the union of all obstacles as $\mathcal{O}$, and the following stage reward function:
$$
\rho(x_k, u_k, x_{k+1}) =
\begin{cases}
-o & \text{if } x_{k+1} \in \mathcal{O}\\
-1 & \text{if } x_{k+1} \notin \mathcal{O} \text{ and } b_k > 0\\
0 & \text{if } x_{k+1} \notin \mathcal{O} \text{ and } b_k = 0
\end{cases}
$$
where $o$ is a positive obstacle collision penalty, which should be taken large so that obstacle avoidance is given priority over minimizing time to transmit. Define also the long-term value function:
\begin{equation}\label{eq:Vh}
V^\pi(x_0) = \sum_{k = 0}^\infty \rho(x_k, u_k, x_{k+1})
\end{equation}
where $x_{k+1} = f(x_k, u_k, R(p_k))$ and $u_k = \pi(x_k)$ obeys the state feedback law $\pi : X \to U$. Here, as well as in the sequel, $p$ is the position component of $x$. Then, the optimal value function is:
\begin{equation}\label{eq:hstar}
\max_\pi V^\pi(x_0) =: V^*(x_0), \ \forall x_0
\end{equation}
and an optimal control law $\pi$ that attains $V^*$ is sought. A choice is made here to (equivalently) use maximization instead of minimization since the learning method to solve PT will be a particular type of reinforcement learning, where value maximization is the convention. Note that the horizon is not known in advance, since the trajectory must run until the buffer is empty. The number of steps until this event occurs depends on the initial buffer size and on the positions along the trajectory. The problem in thus handled in the infinite-horizon setting, per \eqref{eq:Vh}.

When there are no obstacles, $\mathcal{O} = \emptyset$, the optimal solution of \eqref{eq:hstar} will directly minimize the number of steps until the buffer size becomes zero, from any initial state.

When there are obstacles, the method proposed to handle them is very simple. For instance, the robot may still intersect an obstacle in-between two samples, which can be handled by artificially enlarging the obstacles so that even if intersections happen around the edges, a collision does not actually occur. For simplicity, rectangular obstacles are considered; if they have a different shape, a rectangular bounding box may be taken, suitably enlarged as described above. Moreover, because obstacles are handled via penalties in the reward function, the optimal solution changes and is no longer exactly minimizing time. In practice however, when $o$ is large enough, the penalty is expected to act as a constraint that eliminates solutions that intersect obstacles, and the remaining solutions will be close to time-optimal subject to this constraint. To get an idea of how large $o$ should be, consider as an example the case when $b_0 \leq \overline b$ and the rate is lower bounded by some value $\underline{R}$ at every $p$, i.e., $R(p) \geq \underline{R} > 0$, which could represent a minimum quality-of-service requirement. In that case, the buffer will be emptied in at most $\ceil{\overline b/ \underline{R}}$ steps from any initial state, where $\ceil{\cdot}$ denotes ceiling: the smallest integer larger than or equal to the argument. So, any $o > \ceil{\overline b/ \underline{R}}$ will lead to a collision reward that is immediately smaller than any collision-free trajectory value, meaning that the solution will be discarded. Note that $\underline{R} > 0$ is not a requirement for the method to work -- instead, when rates can be zero the penalty can be selected larger than a typical time to transmit the buffer, if one is available from prior knowledge, or simply by trial and error.}

As a necessary first step, \secref{sub:transmit:dp} presents a dynamic programming algorithm to solve PT for known, deterministic rate functions $R$. Afterwards, \secref{sub:transmit:learn} gives the learning-based solution for unknown rate functions; \rev{see again \tabref{tab:roadmap}.}

\subsection{Model-based algorithm for known rate functions}\label{sub:transmit:dp}

If the rates are deterministic and the rate function $R(p)$ is known, dynamic programming (DP) can be applied to PT. Construct an initial value function $V_0(x) = 0, \forall x$, and then iterate for $\ell \geq 0$:\footnote{Subscript $\ell$ in $V_\ell$ denotes the iteration index, whereas the superscripts used earlier denote either the dependence on the policy $\pi$, in $V^\pi$, or the particular case of the optimal policy, in $V^*$.}
\begin{equation}\label{eq:exactDP}
V_{\ell+1}(x) = \max_{u \in U} \left[ \rho(b) + V_\ell( f(x, u, R(p)))\right], \forall x
\end{equation}
where $b$ is the buffer size component of state $x$. Note that knowledge of $R$ is required. The algorithm is stated ``forward in iterations'', but can also intuitively be seen as running ``backwards in time'' as would usually be done in finite-horizon applications of DP.
After stopping the algorithm at some finite iteration $\overline\ell$, the following state feedback is applied:
\begin{equation}\label{eq:policy}
\pi(x) \in \argmax_{u \in U} \left[ \rho(b) + V_{\overline\ell}( f(x, u, R(p)))\right]
\end{equation}
with ties between maximizing actions resolved arbitrarily. A reasonable choice for the number of iterations can be made a priori when rates are always nonzero: $\overline\ell = \ceil{\overline b/ \underline{R}}$, the maximal number of steps to empty the buffer, per the discussion above.

In general, the algorithm is not implementable as given above, for several reasons: the maximization over $u$ is a possibly nonconcave and nondifferentiable global optimization problem, $V$ cannot be exactly represented for continuous arguments $x$, and the rates may be zero at some positions. Below some empirical solutions to these issues are given, which are rather standard in the field of approximate dynamic programming \citep{Bertsekas:12,SuttonBarto:18}. First, $U$ is assumed to consist of a finite, discrete set of actions, and maximization is solved by enumeration. Second, $P$ is assumed bounded and rectangular\footnote{The shape of $P$ can be generalized at the expense of some more intricate interpolation procedure.} and that $b \in [0, \overline{b}]$ (ensured e.g. if $b_0 \leq \overline{b}$), which is reasonable in practice. Value function $V$ is then represented approximately, using multilinear (finite-element) interpolation over grids defined along the interval domains of each of the state variables, per the procedure of \citet{aut10}. Denoting the approximate value function by $\widehat V$, this representation can be written:
\begin{equation}\label{eq:Vhat}
\widehat V(x; \theta) = \transp \varphi(x) \theta
\end{equation}
where $\theta \in \Real^n$, $\varphi: X \to \Real^n$, and $n$ is the total number of points on the grid. Here, $\theta_i$ is the parameter associated with point $i$ and $\varphi_i(x)$ is the weight with which point $i$ participates to the approximation, which is easy to obtain from the interpolation procedure. Note that in fact $\varphi(x)$ will be sparse, $0$ for most $i$; indeed the maximal number of points participating to an interpolated value is $2^{3+n_y}$ (2 dimensions for $p$, $1$ for $b$, and the dimensions of $y$). Note that since the grid size increases exponentially with the number of dimensions, interpolation is feasible only for small $n_y$ (say, up to $3$); for larger dimensions, better function approximators are required, such as neural networks. Still, for rough models of robot motion such a small $n_y$ will often be sufficient.

Noticing that at point $x_i$ of the grid, $V(x_i) = \theta_i$ since the vector $\varphi(x_i)$ is $1$ at position $i$ and $0$ everywhere else, an approximate version of \eqref{eq:exactDP} can be given:
\begin{equation}\label{eq:ADP}
\theta_{\ell+1,i} = \max_{u \in U} \left[ \rho(b) + \widehat V(f(x_i, u, R(p_i)); \theta_\ell)\right], \forall i
\end{equation}
where vector $\theta_0$ is initialized to zero values.

To circumvent the need to fix the number of iterations in advance, the algorithm is stopped when $\norm{\theta_{\ell+1} - \theta_{\ell}}_\infty \leq \eps$. Finally, a control law is computed with an equation similar to \eqref{eq:policy} but using $\widehat V(\cdot; \theta_{\ell+1})$ on the right hand side.

In practice, the accuracy can be increased by making the state interpolation grids and the action discretization finer, and $\eps$ smaller. A discounted version of such an interpolated DP algorithm has been analyzed by \citet{aut10}.

\subsection{Rate-learning algorithm for the transmission problem}\label{sub:transmit:learn}

\begin{figure}[!htb]
  \centering
  \includegraphics[width=0.35\columnwidth]{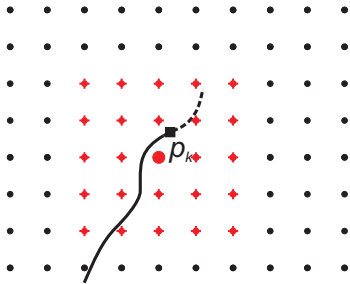}
  \caption{Illustration of local region over which the algorithm performs DP sweeps across the two position axes of the state. The grid is denoted by dots, and the current position of the robot by a square. The center of the subgrid (larger disk) is the nearest grid point below and to the left of the current position, and the subgrid (red crossed circles) extends $r_{\mathrm{DP}}=2$ points in each direction from this center point. The prior trajectory of the robot is the continuous line, and the dashed line illustrates a possible trajectory across several future steps.}\label{fig:subdp}
\end{figure}

Consider now the setting where $R$ is initially fully unknown, and can only be sampled at each position reached by the robot. From the samples $(p_j, R_j)$, $j \leq k$ seen so far, an estimate $\widehat R$ of $R$ is built using any function approximation (supervised learning) technique. Before taking a decision at step $k$, the algorithm uses $\widehat R$ in order to run several DP sweeps of the form \eqref{eq:ADP}, but only \emph{locally}, around state $x_k$, as follows. \rev{A  local subgrid around current state $x_k$ is first selected, consisting of $r_{\mathrm{DP}}$ grid points to either side of the current state along all $3+n_y$ dimensions. To formalize the procedure, define explicitly the discretization grid of each state variable $x^m$ to consist of points $x^m_{i_m}$, $i_m \leq n_m$, so that the total number of grid points is $n = \prod_{m=1}^{3+n_y} n_m$. Then, the algorithm chooses a subgrid center: the point that is closest on the grid and smaller than the current state $x_k$, on each dimension. This point is at index $i_{k,m}$ so that $x^m_{i_m} \leq x^m_k$ but $x^m_{i_m+1} > x^m_k$ (or $i_m = n_m$). Finally, the algorithm selects points at indices $\max\{1, i_{k,m}-r_{\mathrm{DP}}\}, \dotsc, \min\{n_m, i_{k,m}+r_{\mathrm{DP}}\}$ as the subgrid on dimension $m$, where $\max$ and $\min$ enforce the grid bounds. DP sweeps are run by applying \eqref{eq:ADP} $\ell_{\mathrm{DP}}$ times, but only for the points $i$ resulting from the combinations of the per-dimension subgrids defined above. \figref{fig:subdp} illustrates the idea. The DP range $r_{\mathrm{DP}}$, together with the number of DP sweeps $\ell_{\mathrm{DP}}$, are tuning parameters of the algorithm.}

A simple reason for these local updates is to reduce computational costs, since a decision must be made online. A deeper motivation however is to avoid extrapolating too much from the samples of $R$ seen so far, which are all probably behind the robot along its trajectory, and not in the direction that it needs to go; and for the same reason, one cannot hope anyway for a decision that is good across more than a few steps -- i.e., for smooth dynamics, more than a small distance away in the state space. Indeed, it is likely better to wait until more information is available before attempting to construct such a decision.\footnote{This is also a key feature of receding-horizon predictive control, so one may wonder why this framework is not applied here. In fact, a receding-horizon method based on tree search has been attempted, but it performed poorly.}

\rev{When obstacles are present, this method only needs to know about them when one of them contains points reachable from the subgrid. This implies that a sensor with a range on the order of half the subgrid length is sufficient. On the other hand, it also means that in order for the discretization-based DP to work, the grid spacing must be smaller than the smallest obstacle size, and the action discretization and sampling period should be taken such that distances between consecutive positions are also smaller than the safety region with which the obstacle have been enlarged, per the discussion in \secref{sec:problem}. Moreover, $r_{\mathrm{DP}}$ should be large enough so that the stopping distance of the robot is smaller than the half the subgrid length, even when moving at maximal velocity towards the obstacle.}


A remaining question is how exactly the robot chooses actions at each step. The algorithm must \emph{explore} in order to find informative samples, with which to build a better rate function and thus a better value function. This need for exploration is typical in learning problems, and also appears in reinforcement learning, adaptive control, etc. A simple, optimistic exploration strategy is proposed here, based on an optimistic initialization of $V$ (via the parameters $\theta_0$) to some $\overline V$ that is larger than the optimal values. Then the greedy policy will still work well; intuitively, optimistic initialization will force the algorithm to explore the space of solutions since it believes any unseen solution to be good. If the maximal rate $\overline R$ is known, the optimistic initial solution can be obtained by initializing the parameter $\theta_{0,i}$ for each grid point $x_i$ to $- b_i/\overline R$. If $\overline R$ is unknown, then $\theta_0$ should be taken $0$.


Algorithm \ref{alg:PT} summarizes the overall procedure to solve PT. Note that in line \ref{ln:dpupdate}, the rate function approximator $\widehat R$ is used.
\rerev{Compared to the algorithm of \citet{acc19}, this version eliminates two components that did not significantly contribute to performance: direct reinforcement learning and softmax exploration. Instead, here only greedy action selection is used, with optimistic initialization. On the other hand, the algorithm here works for any dynamical order, and implicitly handles obstacles as explained above.}
\begin{algorithm}[!htb]
    \caption{Learning control for PT.} \label{alg:PT}
    \begin{algorithmic}[1]
    \Input $g$, $\overline R$ if known, state grids, discretized actions $U$, $r_{\mathrm{DP}}$ and $\ell_{\mathrm{DP}}$ for DP sweeps
    \State for all grid centers $x_i$: $\theta_{0,i} = -b_i/\overline R$ if $\overline R$ known, $0$ if $\overline R$ unknown
    \State measure initial state $x_0$
    \Repeat{ at each time step $k = 0, 1, 2, \dotsc$}
        \State sample $R_k$, update approximator $\widehat R$
        \State $\aux\theta_0 = \theta_k$, and construct DP subgrid around $x_k$
        \For{DP sweep $\ell=0, \dotsc, \ell_{\mathrm{DP}}-1$}
            \For{each point $i$ on the subgrid}
                \State $\aux\theta_{\ell+1,i} = \max_{u \in U} [ \rho(x_i, u, f(x_i, u, \widehat R(p_i))) + \widehat V(f(x_i, u, \widehat R(p_i)); \aux\theta_\ell)]$
            \label{ln:dpupdate}
            \EndFor
        \EndFor
        \State $\theta_k = \aux\theta_{\ell_{\mathrm{DP}}}$
        \State $\u_k = \argmax_{u \in U} [\rho(x_i, u, f(x_i, u,  r_k)) + \widehat V(f(x_i, u, r_k); \theta_\k)]$
        \State apply action $u_k$, measure next state $x_{k+1}$
    \Until $\b_{k+1} = 0$
\end{algorithmic}
\end{algorithm}

So far, the approximator $\widehat R$ has been left unspecified; again, in principle any sample-based function approximator can be used. For the experiments, local linear regression (LLR) will be used. To this end, each new pair $(p_k, R_k)$ that was not yet seen is stored in a memory. Then, for each query position $p$, the $N$ nearest neighbors of $p$ are found using the Euclidean norm, and linear regression on these neighbors is run to find an affine approximator of the form $\transp \alpha p + \beta$ with $\alpha \in \Real^2, \beta \in \Real$. This approximator is then applied to find $\widehat R(p)$. The tuning parameter of LLR is $N$. An important note is that the robot will sometimes move in straight lines, in which case the positions of the nearest neighbors will be linearly dependent. In that case, when possible, the algorithm selects a smaller set of linearly-independent neighbors. When this is not possible, the approximator directly returns the rate value of the (first) nearest neighbor.

The DP-sweep method is related to model-learning techniques such as Dyna \citep{Sutton:90}, which finds a model from the samples and then applies DP updates to it. 
The method can also be seen as reusing data in-between RL updates, which bears similarities with several classical data reuse techniques from RL. One such technique is experience replay \citep{Lin:92}, which reapplies learning updates to memorized transition and reward samples, and which has recently experienced a resurgence in the field of deep RL \citep{Mnihetal:15}. Nevertheless, the method proposed here is unique due to the specific structure of the problem considered, which allows focusing the learning algorithm on the key unknown element: the rate function.

\section{Empirical study in the transmission problem}\label{sec:transmit:results}

Consider a simulated robot with motion dynamics \eqref{eq:motiondynamics} given by the nonlinear, unicycle-like updates:
\begin{equation}
\begin{aligned}
p_{k+1,1} & = p_{k,1} + T_{s} u_{k,1} \cos(u_{k,2})\\
p_{k+1,2} & = p_{k,2} + T_{s} u_{k,1} \sin(u_{k,2})
\end{aligned}
\end{equation}
i.e., the first input is the velocity and the second the heading of the robot. A set of discretized actions is taken that consists of moving at velocity $u_1 = 1$\,m/s along one of the headings $u_2 = 0, \pi/4, \dotsc, 7\pi/4$\,rad; together with a $0$-velocity action. The sampling period is $T_s = 4$\,s. Note that since these dynamics are first-order, there is no extra motion state $y$, and $x = \concatt{\transp p, b}$. The domain $P = [0, 200]\times[0,200]$\,m, with the bounds enforced by saturation, and $b \in [0, \overline\b] = [0, 1000]$\,Mbit.\footnote{To keep the numbers reasonable, the buffer and bitrates are measured in Mbit and Mbps, respectively.} Such first-order models of mobile robots are useful for high-level control, as they are employed here; usually, an accurate low-level motion controller is applied to implement the reference given by these high-level controllers on the actual robot. Sometimes, even simpler integrator models are used, e.g.\ in consensus theory \citep{OlfatiSaberFaxetal:07}.

The rate function consists of the summation of two path-loss functions of the type in \eqref{eq:Rpath}, initially \emph{deterministic}, with $z = 1$ in \eqref{eq:Rpath}, so that the algorithms for PT work; see e.g.\ Figures \ref{fig:baselines} and \ref{fig:transmit_traj} of \secref{sub:transmit:baselines} for a contour plot of the overall rates. This rate function corresponds to two antennas, used at the same time with dual transmitters by the robot. \rev{The antennas are placed at coordinates $100, 170$ and $100, 30$, respectively, and $K=10^4$, $h=1$, $\gamma=2$ (corresponding to free space) in \eqref{eq:Rpath}, \eqref{eq:SNR}. Moreover, $R_0 \approx 0.753$ for the first, top antenna and $0.188$ for the second, bottom antenna, leading to bitrates of about $10$ and $2.5$\,Mbit/s when the robot is at the positions of the two antennas, respectively. For most of the experiments below, three rectangular obstacles are added, with length $42$ and width $2$, artificially enlarged to $50$ and $10$ respectively to avoid collision in-between sampling times. Two obstacles are vertically oriented and centered on $50, 170$ and $150, 170$ respectively; while the third one is horizontal, centered on $100, 100$, see again \figref{fig:transmit_traj}. The obstacle penalty is $o=100$.

The upcoming simulations are organized as follows. First, \secref{sub:transmit:tuning} evaluates the impact of two important tuning parameters of the learning algorithm: the DP range and the number of nearest neighbors. Then, for a larger number of initial positions, \secref{sub:transmit:baselines} compares the learning algorithm against two baselines: the model-based method and a simple gradient-ascent method that does not require to know the rates. Some representative trajectories are also given. Finally, \secref{sub:transmit:random} verifies how the algorithm handles random rates.

\subsection{Impact of tuning parameters}
\label{sub:transmit:tuning}

In \algoref{alg:PT}, an interpolation grid of $31 \times 31 \times 31$ points is taken for the three state variables (two positions and one buffer size), and the initial position is fixed to $p_0 =  \concatt{10, 170}$, which is closer to the stronger antenna but behind an obstacle. The buffer is initially full, $b_0 = \overline b = 1000$, as that is the most interesting scenario in PT.

Preliminary experiments showed that for the number of iterations $\ell_{\mathrm{DP}}$ in the DP sweeps, any value above $10$ works well; therefore, this value is taken $10$. Next, the range $r_{\mathrm{DP}}$ of the DP sweeps is gradually increased from $1$ to $6$, and the number $N$ of nearest-neighbors which drives the LLR approximation is varied in the set $1, 3, 4, 5, 6$ ($2$ is skipped as in that case the approximator becomes a nearest-neighbor one anyway, so it is equivalent to $N=1$). \figref{fig:transmit_param}, left reports the resulting numbers of steps required to empty the buffer. Note that performance is generally not monotonic in $N$ nor in $r_{\mathrm{DP}}$.

\rerev{A first important observation is that, for this problem, nearest-neighbor approximation works best. This may be due to the relatively simple rate function used. Another reason could be related to the fact that $N>1$ is not beneficial when rate samples are in straight lines, as then the approximator degenerates to nearest-neighbor anyway, as explained in \secref{sub:transmit:learn}. Indeed, this is observed to be the case in 11 to 57\% of the rate function estimates computed by the algorithm, depending on the particular experiment.} Secondly, the buffer is generally emptied in fewer steps for intermediate DP ranges. A likely interpretation is that if the range is too small, the $\widehat R$ learned is not sufficiently exploited by DP; on the other hand, if the range is too large, DP attempts to extrapolate too far from $\widehat R$. The best combination is $r_{\mathrm{DP}} = 4$, $N = 1$, and these values will be used in all the upcoming experiments.
\begin{figure}[!b]
  \centering
  \includegraphics[width=0.48\columnwidth]{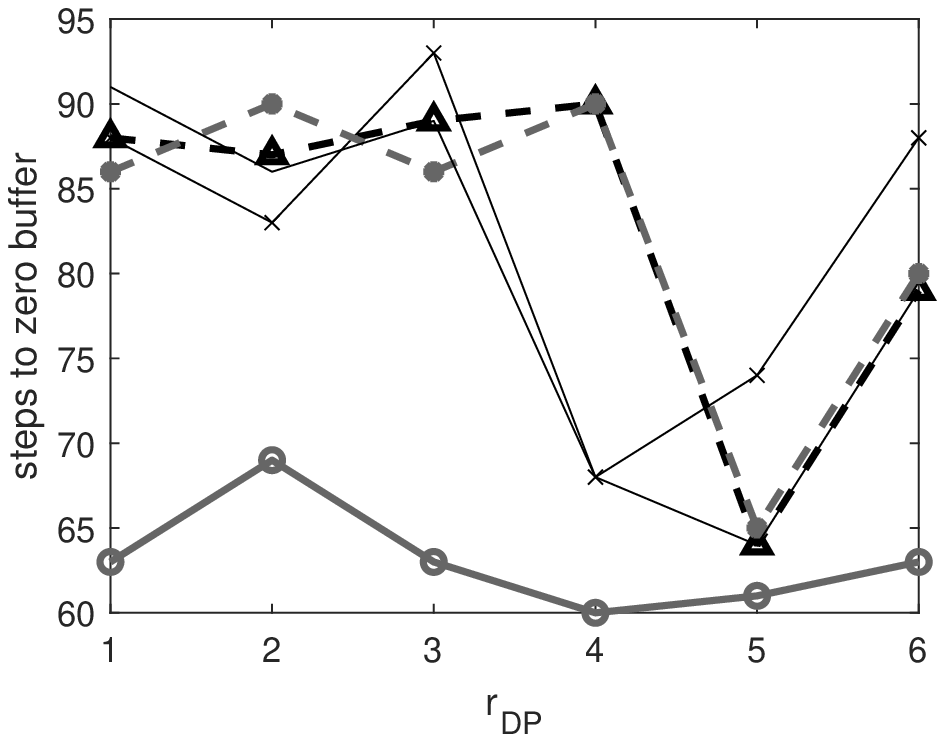}
  \includegraphics[width=0.48\columnwidth]{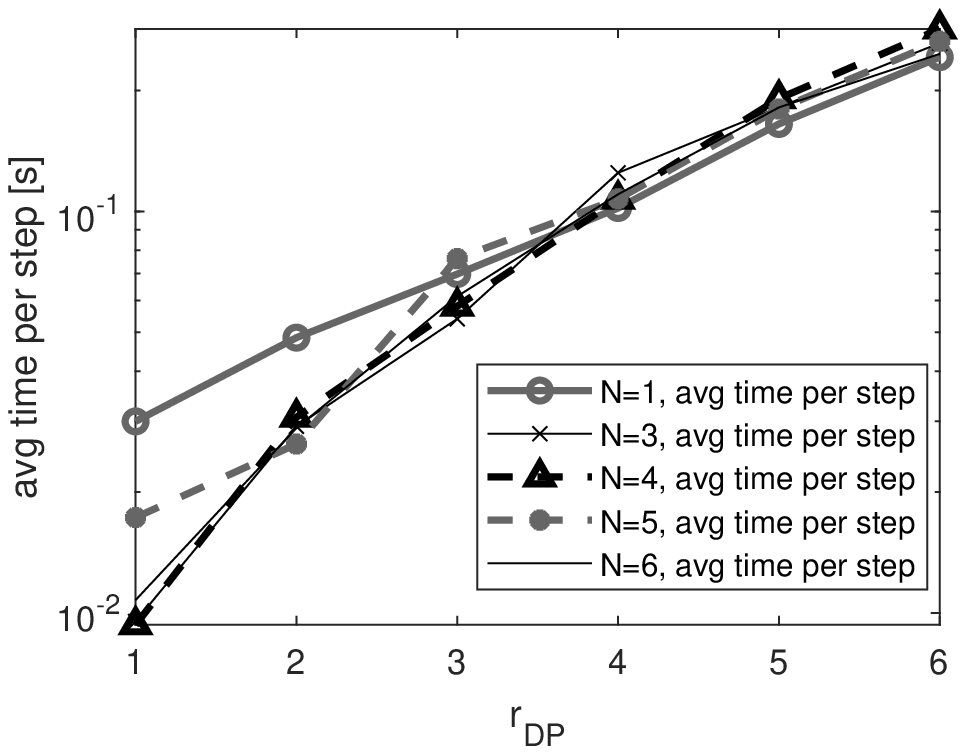}
  \caption{\rev{Influence of the DP range and number of nearest neighbors in PT. Each curve is for one value of $N$, see the legend of the figure on the right.}}
  \label{fig:transmit_param}
\end{figure}

Consider next computation time, shown in \figref{fig:transmit_param}, right. \rerev{Increasing the DP range requires of course larger computational costs, here roughly cubical in $r_{\mathrm{DP}}$ due to the three-dimensional state space. This effect dominates computation time for the scenario considered, and the impact of the number of nearest neighbors is small, as seen from the fact that the curves are very similar for large $r_{\mathrm{DP}}$.} Overall, for most settings a control frequency of about $10$\,Hz can be achieved, which is adequate for high-level control of the robot position.

\subsection{Comparison to baselines}
\label{sub:transmit:baselines}

A first, natural baseline to compare against is the model-based method of \secref{sub:transmit:dp}. The same  $31 \times 31 \times 31$ interpolation grid is taken as above. The model-based method and the learning \algoref{alg:PT} are run from $18$ initial positions selected uniformly randomly in the domain $P$, always with a full initial buffer, $b_0 = \overline b = 1000$. \figref{fig:baselines}, left shows the results. With the exception of a few outliers, learning generally empties the buffer in less than twice the number of steps taken by the model-based method. This is a good result, keeping in mind that the rate function must be learned \emph{at the same time as} using it to transmit. Note that there are some lucky initial states in which the learning algorithm is faster; since the model-based method also has approximation errors, it is not guaranteed to always produce a better solution.

\begin{figure}[!htb]
  \centering
  \includegraphics[height=6.5cm]{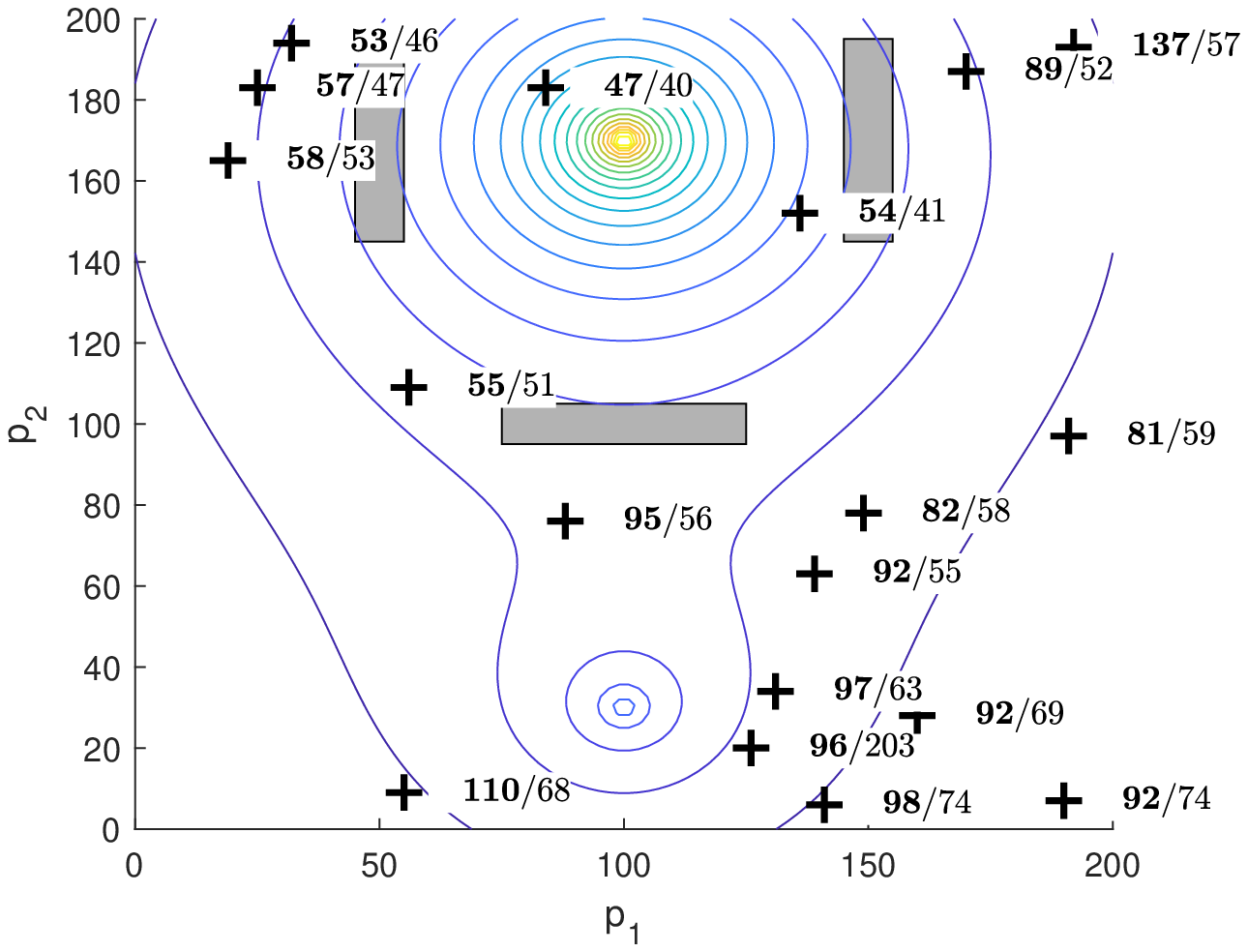}\hspace*{0.2cm}
  \includegraphics[height=6.5cm]{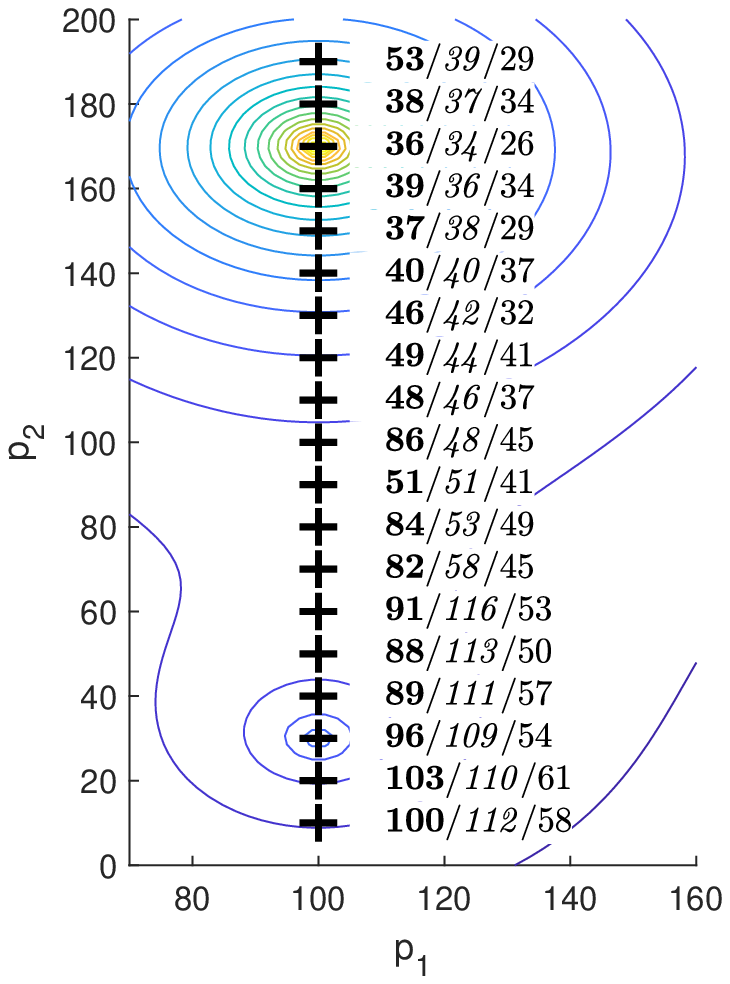}
  \caption{\rev{Comparison of \algoref{alg:PT} with model-based control (left); and with gradient ascent and model-based control in an obstacle-free scenario (right). Each cross denotes a starting position, labeled by the number of steps to zero buffer for the learning algorithm (in bold), for the model-based method (in regular font), and for the gradient ascent baseline (in italic). For readability, the horizontal position is restricted to the interval $[70, 160]$ in the graph on the right.}}
  \label{fig:baselines}
\end{figure}

It will also be instructive to compare \algoref{alg:PT} to a baseline that does not require knowledge of the rate function. A ``myopic'' baseline method is proposed here, which performs gradient ascent on the estimated rate function in the following way. At each discrete time step, LLR is performed around the current state; thus, rate function learning is performed in the same way as in \algoref{alg:PT}. The plane produced by LLR is differentiated to provide a local estimated gradient of the rate with respect to the position, and the robot moves with unit velocity in the direction of this gradient. Note that $N$ in LLR must be selected at least $3$ to produce a plane. When the nearest neighbors are not linearly independent, the gradient cannot be found. Instead, to preserve repeatability, the method then chooses deterministically a direction to move in, which rotates with the step: $0$\,rad at $k=0$, $\pi/2$ at $k=1$, etc.

The gradient method cannot perform obstacle avoidance, so to compare with it, the obstacles are removed. Moreover, the gradient method will not produce very different results from \algoref{alg:PT} in the ``basin of attraction'' of the strong antenna, which includes many initial states. So, to have an interesting (and still fair) comparison, instead of random positions a set of equidistant positions are selected, spaced along the line defined by the two antenna centers, between $10$ and $190$ with a step of $10$. For the gradient method, $N=3$ nearest neighbors are taken in LLR. \figref{fig:baselines}, right shows the results, comparing not only to \algoref{alg:PT}, but also to model-based control. As predicted, when initialized around the stronger antenna, the gradient method performs quite well; in such states, although learning is not far behind, it must pay a price for the extra generality of handling arbitrarily-shaped rate functions. However, for states around the weaker antenna, where it is nevertheless still long-term optimal to travel to the stronger antenna, the non-myopic, DP-based \algoref{alg:PT} does better. This advantage will be important in more realistic problems where the rate function shape is more complicated, making non-myopic decisions important.

To get more insight, consider next representative trajectories of the model-based algorithm and of \algoref{alg:PT}, in the scenario with obstacles. The initial position is $p_0 = \concatt{40, 150}$. \figref{fig:transmit_traj} gives the results. In contrast to the model-based solution, which goes along the shortest path around the obstacle towards an antenna (since it knows where it is), the learning algorithm must explore a less direct path to observe samples from $R$ and build its estimate. A number of $64$ steps are required to empty buffer, which is about $30\%$ worse than the model-based performance ($49$ steps).
\begin{figure}[!htb]
  \centering
  \includegraphics[width=0.49\columnwidth]{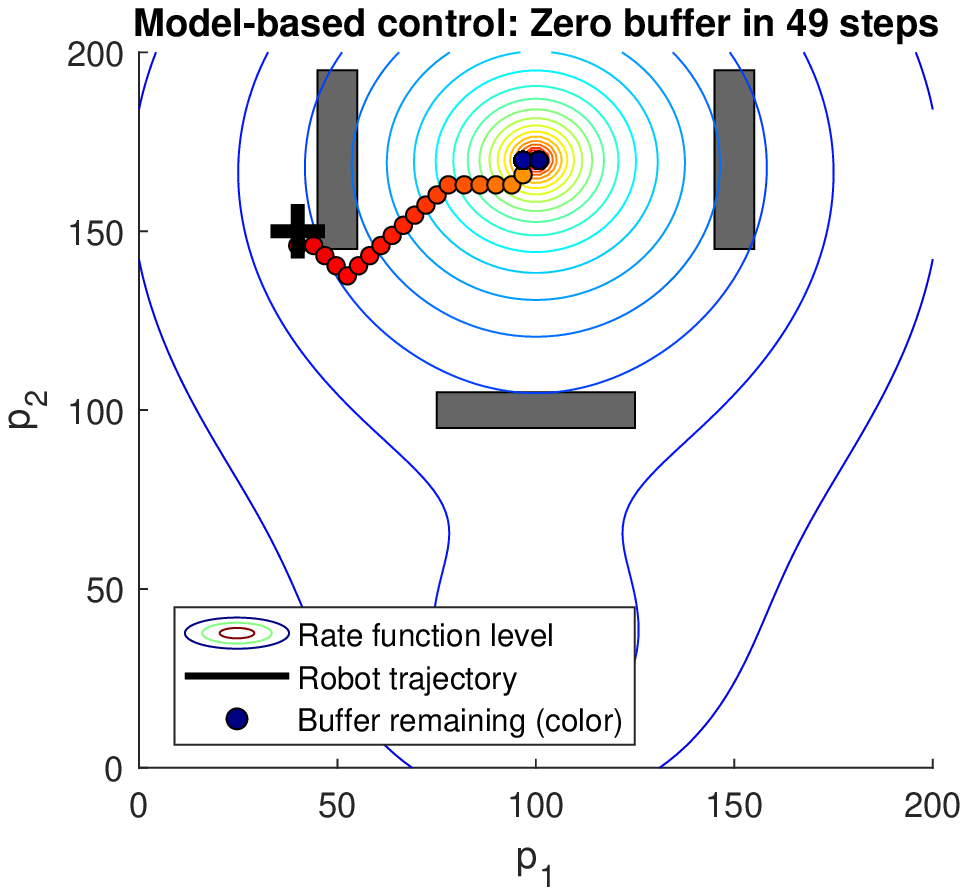}
  \includegraphics[width=0.49\columnwidth]{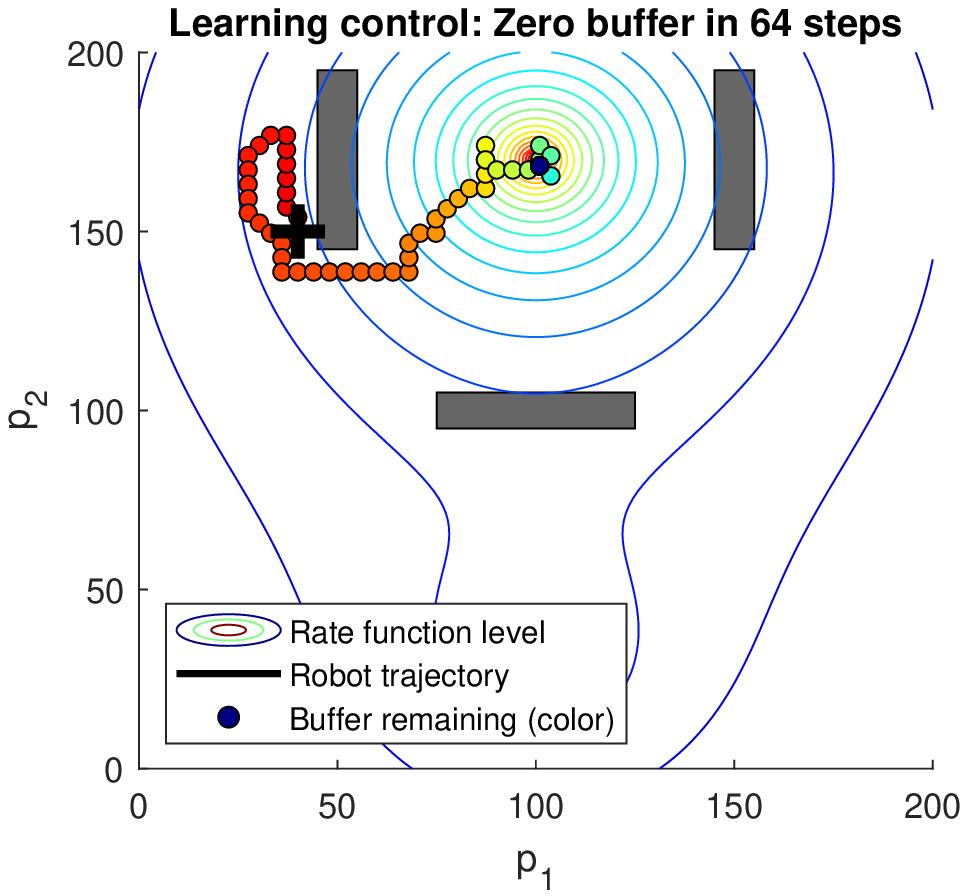}
  \caption{\rev{Trajectories with model-based control (left) versus learning control (right) in PT. The contour plot shows the rate function, and gray rectangles are (enlarged) obstacles. The position of the robot at each sampling time is shown by a colored disk. The color of the disk indicates the remaining buffer size, from dark red (full) to dark blue (empty). The black ``+'' marks the starting position of the robot.}} 
  \label{fig:transmit_traj}
\end{figure}
}

\subsection{Impact of random fluctuations}
\label{sub:transmit:random}

\begin{figure}[!htb]
  \centering
  \includegraphics[width=0.48\columnwidth]{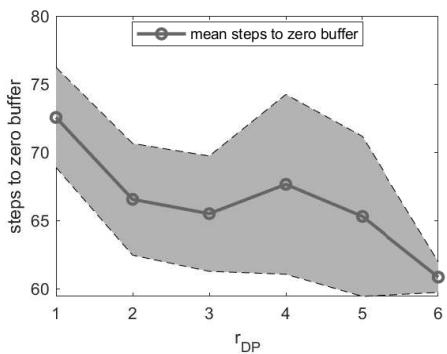}
  \caption{\rev{Influence of the DP range for random rates in PT. The curve with markers shows the mean number of steps to empty the buffer, and the shaded regions give the $95$\,\% confidence interval on this mean.}}
  \label{fig:transmit_noisy}
\end{figure}

Before finishing the simulations in PT, additional experiments are perofmed to check how the algorithm handles random rates, even though it was not explicitly designed for this. In particular, Rice fluctuations $z$ with $v = 15$ are introduced in the SNRs of both antennas, leading to a moderate variance: most of the probability mass is assigned to $z \in [0.7, 1.3]$ so the SNR can vary by about $30$\% either way. The initial position is the same as in \secref{sub:transmit:tuning}. \figref{fig:transmit_noisy} reports results for varying $r_{\mathrm{DP}}$ with fixed $N=1$. For each value of $r_{\mathrm{DP}}$, $20$ independent runs of the algorithm are performed, and the mean performance is reported, together with the $95\%$ confidence interval on this mean. \rev{Interestingly, in this case a large $r_{\mathrm{DP}}$ seems to work better, in which case performance is very close to that in the deterministic-case.} Overall, the algorithm still works reasonably well in the presence of random fluctuations.

\section{Solution for the navigation and transmission problem}\label{sec:nav}

As for PT, first provides a model-based procedure, in \secref{sub:nav:mb}. This procedure is not a contribution of this paper, but is adapted from \citet{Loheacetal:19}. Then, \secref{sub:nav:learn} gives the learning procedure for unknown SNR functions, which is a novel contribution; see again \tabref{tab:roadmap}.

\subsection{Model-based algorithm for known rate functions}\label{sub:nav:mb}

To start with, the motion dynamics are rewritten as $p_{k+1}=p_k+T_s u_k$, with $n_u=2$ inputs; this is always possible for the simple-integrator motion dynamics (with $n_y=0$) that was assumed in PN. Further, it is assumed that the set of admissible controls $U$ is the closed unit ball of $\Real^2$. The continuous time version of the navigation problem (PN) is to find --- given and initial position $p^0$, an initial buffer $b^0\in\Real_+$ and a target position $p^*\in\Real^2$ --- the minimal time $T\geqslant0$ such that there exist a control $u\in L^\infty(0,T)^2$, satisfying $\|u(t)\|\leqslant 1$ for almost every $t\in (0,T)$, such that:
$$p(T)=p^*\quad\text{and}\quad b(T)\leqslant 0,$$
where $p$ and $b$ are solution of
\begin{align*}
	\dot{p}& = u & p(0)=p^0,\\
	\dot{b}& = R(p) & b(0)=b^0.
\end{align*}
Given a change of time variable and a rescaling of $R$, the control constraint $\|u(t)\|\leqslant 1$ is not restrictive. In fact, this case covers all the constrained inputs of the form $\|u(t)\|\leqslant M$ for every $M>0$. In the absence of fluctuations, the transmission rate $R$ is only a function of the distance between the robot and the antenna, i.e.,
$$R(p)=C(\|p-p_{\mathrm{ant}}\|)\qquad (p\in\Real^2),$$
where $p_{\mathrm{ant}}$ is the position of the wireless antenna.

This time-optimal control problem has been studied by~\citet{Loheacetal:19}.
Using the Pontryagin maximum principle, the optimal controls have been characterized, and the results are briefly summarized here.
To this end, define the map $B$ by:
\begin{equation}\label{eq:B}
	B(q^0,q^1)
	=\|q^1-q^0\|\int_0^1 C(\|q^0-p_{\mathrm{ant}}+s(q^1-q^0)\|)\, \d s\qquad (q^0,q^1\in\Real^2).
\end{equation}
This quantity represents the amount of buffer transmitted when going in a straight line from $q^0$ to $q^1$ with velocity one.
Note also that in order to obtain the following results, $C$ must be an absolutely continuous and nonnegative function defined on $\Real_+$, and must be strictly decreasing on the set where it is nonzero.
\begin{enumerate}
	\item When the initial buffer is small, i.e. $b^0\leqslant B(p^0,p^*)$, then the optimal robot path is to go straight to the target $p^*$ with maximal velocity.
	\item When the buffer is large, i.e., $b^0\geqslant B(p^0,p_{\mathrm{ant}})+B(p_{\mathrm{ant}},p^*)$, then the optimal robot path first goes straight to the antenna, waits at the antenna for a certain amount of time, and finally goes straight to the target.
	\item When the initial buffer is intermediate, i.e. $B(p^0,p^*)\leqslant b^0\leqslant B(p^0,p_{\mathrm{ant}})+B(p_{\mathrm{ant}},p^*)$, the optimal robot path will never reach the antenna's position, and it is a $C^1$ curve contained in the triangle formed by $p_0$, $p_{\mathrm{ant}}$ and $p^*$.
\end{enumerate}
In the two first cases, the minimal time $T$ and the optimal control $u$ can be explicitly computed, while in the last case, it has been shown that the optimal control is expressed through a parameter $(\epsilon,\eta)$ belonging to a set of the form $\{-1,1\}\times \Omega$, where $\Omega$ is a compact set in $\Real^2$.
Unfortunately, in the last case, an explicit formula for the time optimal control is not available, but this control can be numerically approximated.
Indeed, for $\epsilon\in\{-1,1\}$, the remaining part $\eta\in\Omega$ of the parameter is the root of a certain continuous function $\phi_\epsilon$, and monotonicity properties on $\phi$ allows us, for instance, to use a dichotomy based algorithm, see~\citet{Loheacetal:19} for more details.
Note also that the velocity of the robot is always maximal, except temporarily in case 2, when the robot stops at the antenna position.

To obtain a (heuristic) closed-loop control for the discrete-time setting of the present paper, at each step $k$, the model-based method above is run with $p^0=p_k$ and $b^0=b_k$, for the underlying deterministic rate function.
From the optimal solution obtained in this way, the heading $\alpha_k$ for the robot is chosen as the angle of the optimal control obtained at the  initial time. Then, the robot is driven along this heading with unit velocity, for the duration of the sampling period, and the procedure is repeated at the next step.

\subsection{SNR-learning algorithm for the navigation problem}\label{sub:nav:learn}


Consider next the case where only the form \eqref{eq:SNR} of the SNR function is known, and all or some of the parameters in this form are unknown. In particular, the position of the antenna will generally be unknown.

The first step is to design a learning procedure for the rate function from the data accumulated so far along the trajectory of the robot. Then, once this procedure is in place, the problem of how to control the trajectory will be tackled.

To learn the rate function, one could still use a generic function approximator, like in \secref{sub:nav:learn}. This would however not be a good solution in this case, for two reasons. Firstly, here significant insight into the structure of the rate function is available: it obeys \eqref{eq:Rpath}, with the SNR \eqref{eq:SNR}. \rev{Secondly, a generic approximator would in general not guarantee that rates radially decrease around the antenna, making it impossible to apply the procedure from \secref{sub:nav:mb} since a good function $C$ could not be defined.}

Instead, here the SNR will be learned directly in the form \eqref{eq:SNR} from the database of samples seen so far: $(p_j, S_j)$, $j \leq k$. To this end, define an SNR approximator:
\begin{equation}\label{eq:SNRhat}
\frac{\widehat K}{(\norm{p_k - \widehat p_{\mathrm{ant}}}+\widehat h)^{\widehat \gamma}}
\end{equation}
where the ``hat'' versions of the parameters may either be estimated from data, or known in advance: which is the case for one particular parameter depends on the amount of prior knowledge available. For instance, it may be realistic to know the path loss exponent $\gamma$, since tables of values have been found experimentally depending on the type of environment \citep{Mirandaetal:13}. Since $K$ is largely driven by transmission power, it may also be known. However, the antenna position $p_{\mathrm{ant}}$ and the shape of the SNR, given by $h$, are less likely to be known. Let $W \in \Real^{n_W}$ denote the vector of unknown parameters, and let $\widehat S(p; W)$ denote the value of \eqref{eq:SNRhat} for a given vector $W$. Note that the (deterministic) approximate rate function is therefore:
\begin{equation}\label{eq:RSNR}
\widehat R(p; W) = R_0 \log_2 [1 + \widehat S(p; W)]
\end{equation}

Another key difference between the PN setting here and PT above is that here the SNR is natively random. This is of course more realistic, but another reason for allowing stochastic SNR is that the deterministic case would render the learning problem trivial: usually one would only need to observe as many samples as there are unknown parameters, and then solve the system of (nonlinear) equations $\widehat S(p_j) = S_j$ to get exact values for these parameters.

Instead, given the samples $S_j = z_j S(p_j)$ affected by Rice fluctuations $z$, nonlinear regression is performed to minimize the following mean squared error between the estimated SNR and the samples:
\begin{equation}\label{eq:nlreg}
W_{k} = \argmin_W \frac{1}{k} \sum_{j=1}^k \vert \widehat S(p; W) - S_j \vert^2
\end{equation}
Note that this is done at each step, after every newly observed SNR sample. Preliminary experiments showed that gradient-based algorithms are working poorly, since for larger distances from the antenna the magnitude of the gradients is exceedingly small and drowned by the fluctuations. Instead, gradient-free algorithms are suggested, and for the experiments below Nelder-Mead optimization is used. At each step $k$ except the first, the optimizer is initialized with the previously found estimate $W_{k-1}$.

Consider next the problem of generating control actions for the robot. These actions have two partly conflicting goals: obtaining informative samples of the SNR so as to find better parameter estimates, and using the current estimates to drive the robot along a good trajectory. This is yet another instantiation of the classical exploration-exploitation dilemma.

\textbf{Exploration} is handled by an active learning procedure which drives the robot to points that are more likely to provide new information about the SNR. The starting point is the method of \citet{Wuetal:19}. Just like for PT, $U$ is assumed to be a finite, discrete set of actions. Denote by $P^+_k$ the discrete set of positions reachable in one step from $x_k$, from the current state:
\begin{equation}\label{eq:Pplus}
P^+_k = \cset{p}{p \text{ is the position part of state } g(x_k, u), u \in U}
\end{equation}
Then, define the following quantity for each candidate position $p^+ \in P^+_k$:
$$
d_{\mathrm{inf}}(p^+) = \min_{j \leq k} \norm{p_j - p^+} \cdot \norm{S_j - \widehat S(p^+; W_k)}
$$
Quantity $d_{\mathrm{inf}}$ heuristically indicates how informative $p^+$ is, both in terms of positions via the term $\norm{p_j - p^+}$, which is larger when the candidate position is further away from the previously seen samples, and in terms of SNRs via the term $\norm{S_j - \widehat S(p^+; W_k)}$, which is larger when the SNR is predicted to be more different from the previously seen SNR samples. The so-called improved greedy sampling strategy of \citet{Wuetal:19} chooses to move to a position that maximizes this information indicator:
$$
\argmax_{p^+ \in P^+_k} d_{\mathrm{inf}}(p^+)
$$

In the case considered here, this strategy will not be applied directly, since it would only explore without considering the control objective of PN. Instead, exploration must be balanced with exploitation of the current solution. For \textbf{exploitation} of the robot's current knowledge about the rates, the optimal control procedure of \secref{sub:nav:mb} is used to design at state $x_k$ a heading $\alpha_k$, \emph{with the current estimate} $W_k$ of the parameters. Of course, this heading is not optimal, since the estimates are inaccurate, especially early on during learning. This is addressed by taking an action that in addition to $\alpha_k$ also considers the information indicator $d_{\mathrm{inf}}$ above, thereby balancing exploration with exploitation.

In particular, define for each $p^+ \in P^+_k$ a ``control distance'' $d_{\mathrm{ctl}}(p^+) = \floor{\alpha_k - \alpha(p^+)}$, where $\alpha(p^+)$ is the angle of the vector $(p_k, p^+)$ and $\floor{\cdot}$ denotes the difference between the two angles computed not by a simple absolute value, but instead by taking the shortest route around the circle. Then, the algorithm will choose a control that takes the robot to a position:
\begin{equation}\label{eq:pchoice}
p_{k+1} \in \argmax_{p^+ \in P^+_k} \frac{d_{\mathrm{inf}}(p^+)}{d_{\mathrm{ctl}}(p^+)}
\end{equation}
with ties broken arbitrarily. In this way, headings that are closer to $\alpha_k$ are ranked higher.

Algorithm \ref{alg:PN} summarizes the overall procedure to solve PN. Note that once the buffer is empty, it makes no sense to continue exploring and learning, so instead the algorithm simply computes a heading that takes it to the goal state and moves there at maximum velocity (line \ref{ln:navtogoal}). Moreover, to choose actions in line \ref{ln:usel}, the algorithm needs to store which action generated point $p_{k+1}$.
\begin{algorithm}[!htb]
    \caption{Learning control for PN.} \label{alg:PN}
    \begin{algorithmic}[1]
    \Input $g$, discretized actions $U$, optimizer for regression, known parameters in $S$
    \State initialize unknown SNR parameters $W_{-1}$
    \State measure initial state $x_0$
    \Repeat{ at each time step $k = 0, 1, 2, \dotsc$}
        \State measure $S_k$, add $(p_k, S_k)$ to the database of points
        \State run nonlinear regression \eqref{eq:nlreg} from initial solution $W_{k-1}$, obtaining $W_k$
        \State compute set $P^+_k$ of candidate positions from \eqref{eq:Pplus}
        \State design optimal heading $\alpha_k$ for rate $\widehat R(p; W_k)$  \eqref{eq:RSNR}
        \State choose point $p_{k+1}$ with \eqref{eq:pchoice}
        \State apply action $u_k$ that takes the robot to $\p_{k+1}$ \label{ln:usel}
    \Until $\b_{k+1} = 0$
    \State if $p_{k+1} \neq p^*$, move directly to $p^*$ \label{ln:navtogoal}
\end{algorithmic}
\end{algorithm}

The technique used for active learning above is very simple, and well-suited for real-time control. Many classical active learning methods for regression are rather intricate, probabilistic, and they require developing an ensemble of models in order to e.g.\ build a probability distribution over the model outputs, or to perform voting in the ensemble in the query-by-committee approach. Nevertheless, from the recent works of \citet{Wuetal:19} and \citet{ONeilletal:17} it appears that very simple techniques that rank candidate samples simply by their distances to the existing data points are either on par with, or outperform the probabilistic techniques in extensive experimental studies on standard datasets. Here, the greedy sampling technique of \citet{Wuetal:19} was selected, and extended to take into account the control objective (exploitation). 

\section{Empirical study in the navigation and transmission problem}\label{sec:navresults}

In PN, only one antenna is allowed. It is placed at coordinates $100, 30$, with the same shape as in the PT experiments above, and with $R_0 = 0.753$. The same robot motion dynamics, position domain, and discretized actions are used as before. However, here the rate function is natively random: in all experiments, $z$ in \eqref{eq:Rpath} is random with a Rice distribution, per \eqref{eq:rice}. To account for this randomness, the results always report the mean number of steps to reach the goal, together with the $95$\,\% confidence interval on this mean.

\rev{
Next, \secref{sub:nav:magnitude} studies the effect on performance of the magnitude of the random fluctuations. Then, \secref{sub:nav:baselines} compares learning to a simple baseline that exploits the gradient-ascent idea from \secref{sub:transmit:baselines}.
}

\subsection{Impact of the magnitude of random fluctuations}
\label{sub:nav:magnitude}

To apply \algoref{alg:PN}, parameters $K$ and $\gamma$ of the SNR are considered known, and the algorithm only learns the position $p_{\mathrm{ant}}$ of the antenna and the shape parameter $h$. To find these unknown parameters, a variant of Nelder-Mead optimization is applied that uses bound constraints to ensure that the antenna center is in $P$, and that $h$ is positive \citep{derrico:12}.\footnote{The following settings are used for this Nelder-Mead implementation in Matlab: \texttt{TolFun}=0.1, \texttt{MaxIter}=5000, \texttt{MaxFunEval}=10000.} The initial values in $W_{-1}$ are taken $0, 0$ for the antenna coordinates, and $5$ for $h$.

Two initial states are considered: one with a full buffer as in PT, $x_0^1 = \concatt{30, 140, 1000}$, and another where the buffer is only a quarter full: $x_0^2 = \concatt{30, 140, 250}$. The reason is that from $x_0^1$ the buffer is large enough that (even in the case when the rate or SNR function is fully known) the robot should pass through the antenna position along its way to the goal; $x_0^2$ provides a more interesting scenario where the robot does not have to reach the antenna, and can drive in a curved line towards the goal while still emptying the buffer. 

\begin{figure}[!htb]
  \centering
  \includegraphics[width=0.48\columnwidth]{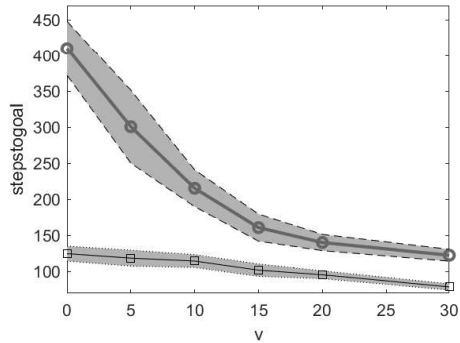}
  \caption{Influence of the SNR variance in PN: $x_0^1$ with thick line and circle markers, $x_0^2$ with thin line and square markers.}
  \label{fig:navigate_param}
\end{figure}

The influence of the variance in the random fluctuations is examined, by gradually changing $v$ in the sequence $0, 5, 10, 15, 20, 30$. Recall that a smaller $v$ leads to larger variance; e.g., for $v=0$, the random variable $z$ can change the SNR by a factor of $4$ with large probability. \figref{fig:navigate_param} reports the resulting numbers of steps required to reach the goal with zero buffer, for the two initial states considered, from $30$ independent experiments. Clearly, larger variance is more challenging, and the effect is greater for the large-buffer initial state.
%

Regarding execution time, it is up to about $0.03$\,s per step on average, so again sufficient for real-time control. Another interesting observation is that if the variance is not too large, the algorithm will most often find good estimates of the antenna position, but it more easily makes mistakes in the shape parameter $h$.

\begin{figure}[!htb]
  \centering
  \includegraphics[width=0.48\columnwidth]{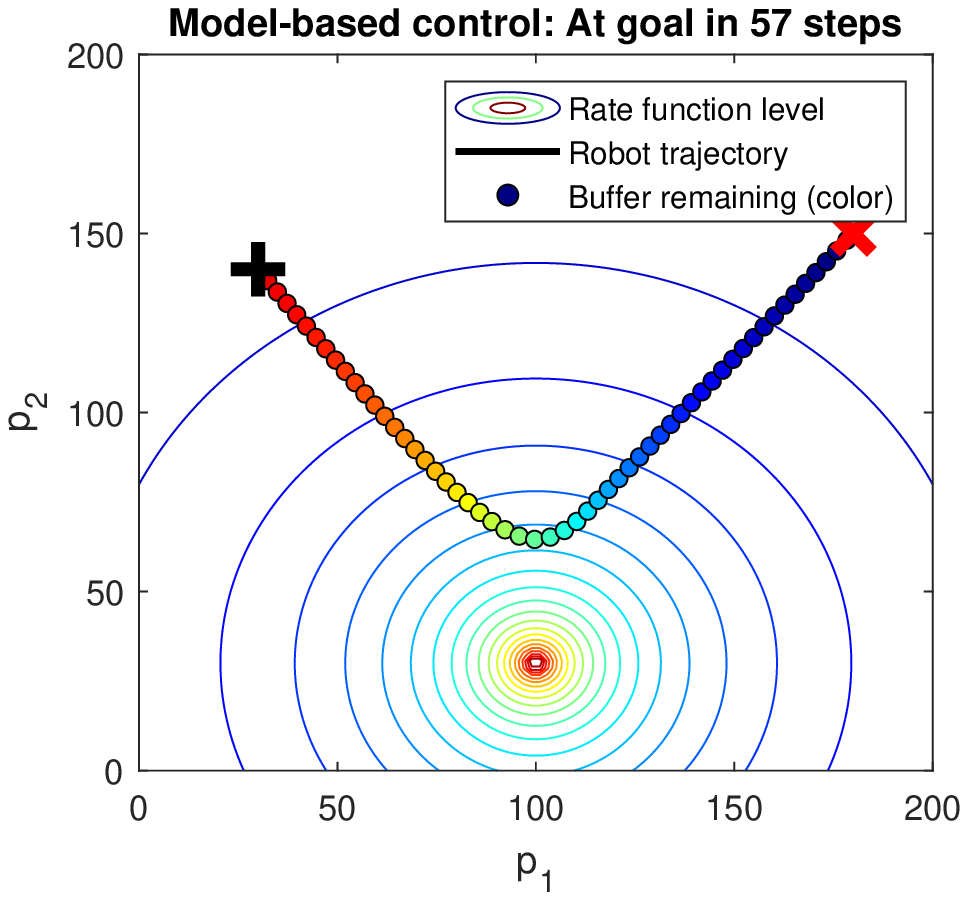}
  \includegraphics[width=0.48\columnwidth]{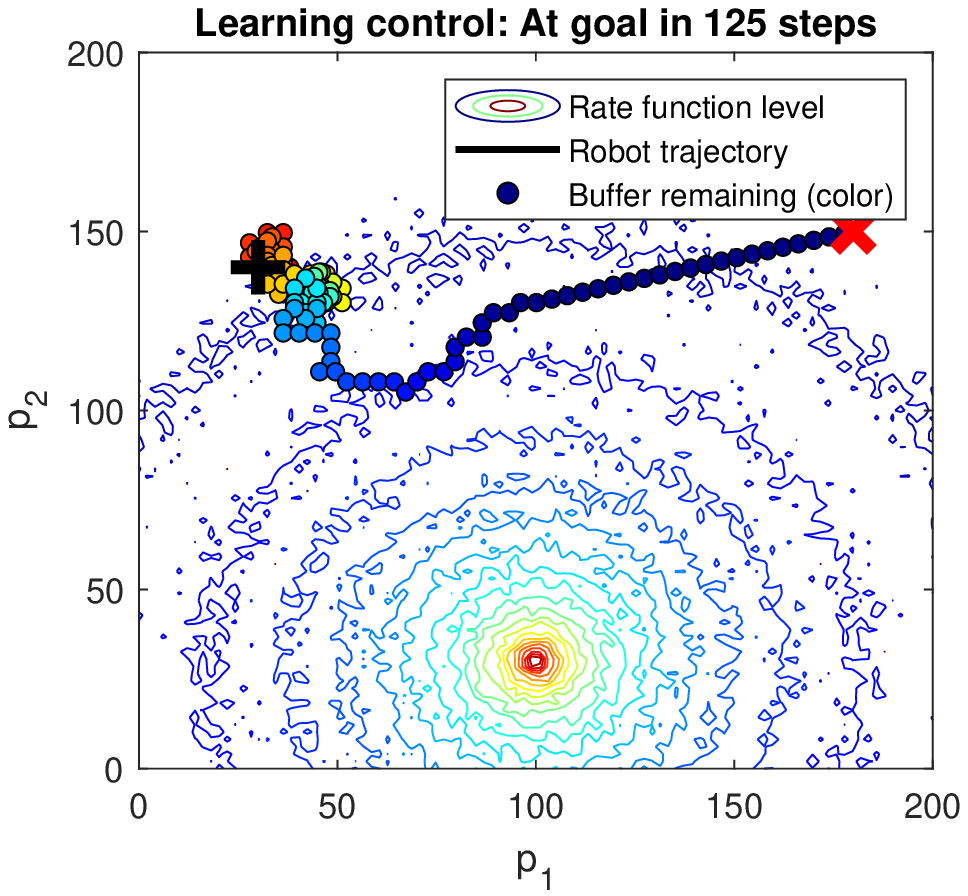}
  \caption{Left: Model-based control (left) versus learning control (right) in PN. The red ``X'' shows the goal location.}
  \label{fig:navigate_traj}
\end{figure}

\rev{
\subsection{Comparison to baseline}
\label{sub:nav:baselines}

To create a simple baseline method, the gradient-ascent procedure of \secref{sub:transmit:baselines} is applied as long as the buffer is nonzero, ignoring the presence of the goal state. Once the buffer has been emptied, the robot goes directly to the goal.  This baseline method is run alongside \algoref{alg:PN} for a selection of $10$ initial states where the positions are taken uniformly random over $P$ (fewer positions than in \secref{sub:nav:baselines} to maintain readability), and the buffer is $b_0 = 250$. Parameter $v$ in the SNR is 15, leading to moderately-sized random fluctuations. \figref{fig:navbaseline} shows the results from $30$ independent runs of both algorithms for each initial state. For all the initial states, the \algoref{alg:PN} is better.

\begin{figure}[!htb]
  \centering
  \includegraphics[width=0.6\columnwidth]{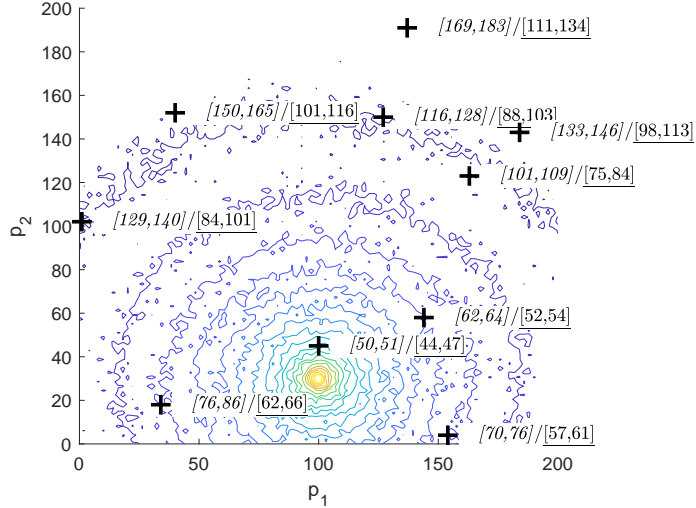}
  \caption{Comparison of PN \algoref{alg:PN} with the gradient-ascent baseline. Each cross denotes a starting position. The italic interval is the $95$\% confidence interval on the mean number of steps to reach the goal for the gradient-ascent baseline method; and the underlined confidence interval is for the learning algorithm. Note that to illustrate the random fluctuations, the contour plot shows a sampled version of the rate function, similar to what the robot would observe.} 
  \label{fig:navbaseline}
\end{figure}

Finally, like for PT in \secref{sub:transmit:baselines}, a trajectory with the model-based method of \secref{sub:nav:mb} is compared with a trajectory with the learning algorithm, for initial state $x_0^2 = \concatt{30, 140, 250}$.} Recall that for the model-based method, the SNR function must be fully known and deterministic. The result with this method is shown in \figref{fig:navigate_traj}, left. The goal is reached in $57$ steps. Note that, due to errors introduced by applying the continuous-time solution of \citet{Loheacetal:19} in discrete-time, the buffer is not emptied exactly as the goal is reached, but a few steps before, so this solution is not fully optimal; still, it is expected to be nearly optimal. A representative trajectory of the learning algorithm, for $v=15$, is shown in \figref{fig:navigate_traj}, right. The learning algorithm explores to find samples of $S$, and once that is done, the robot starts following a trajectory similar in shape to the model-based one, which turns towards the goal. A number of $125$ steps are needed to reach the goal.

\pagebreak[4]
\section{Real-life illustration of the transmission problem}\label{sec:realtime}

An illustration of the transmission problem, PT, is provided for a real quadcopter drone in an indoor environment. Specifically, a Parrot AR.Drone \citep{Bristeau:11}, version 2 will be used, along with a 4-camera OptiTrack Flex 13 motion capture system \citep{Furtadoetal:19}. The high-level motion dynamics \eqref{eq:motiondynamics} used in the learning algorithm have a simple-integrator form (without any extra signal $y$):
$$
p_{k+1} = p_k + T_s u_k
$$
with a sampling period $T_s$ of $6$\,s. There are five discretized controls: one keeps the drone stationary, and the other four move it along one of the cardinal headings $-\pi/2, 0, \pi/2, \pi$, by a distance of $0.3$\,m during a sampling period. This high-level ``virtual'' control $u_k$ is then implemented in reality by generating references that are tracked with lower-level controllers, as it will be explained below. Small steps are taken with the drone so that (i) the lower-level linear control works well despite its limitations and (ii) the drone stays in the region where the position feedback from the OptiTrack system is accurate. In particular, for (ii) the domain $P$ is [-1.2, 1.2] by [-1.2, 1.2]\,m. Thus, ideally the drone moves on an equidistant grid with $7\times 7$ points and a spacing of $0.3$\,m on each dimension. This grid is used for several purposes below.

\rev{Regarding communication, the SNR is sampled offline on the grid, for a fixed transmission power, and interpolated bilinearly between the grid points. Then, the buffer dynamics are simulated using \eqref{eq:bufferdyn}, so that an idealized (but still realistic) transmission rate is obtained from the measured SNR, considering a channel with 20 MHz bandwidth. The main reason for simulating the buffer instead of using real rates is that the experiment must be done in a small area of $2.4$\,m x $2.4$\,m, in which the 4-camera OptiTrack system can provide accurate positioning. In this small area, the SNR values are good, and the WiFi protocol is able to maintain a nearly constant transmission rate. Thus, when using real rates there would be no change of the measured transmission rate with the relative position to the Wi-Fi access point, and the experiment would be uninformative. In a real-life scenario, the drone would have to cover a wide area, transmission rates would naturally vary significantly with the position, and these limitations would not apply. Another consideration in a practical application is that control and sensing values must be sent over the same network as the buffer data. This can be solved e.g.\ through a networked control co-design method, by using a QoS algorithm to dynamically prioritize the closed loop traffic as needed. Note that by relying on the SNR the results are also more generic, while using the rates directly would make the result dependent on device and protocol details (such as the retransmission protocol, interleaving size, modulation scheme, mapping, error correction coder, etc.)}

Interpolation in \algoref{alg:PT} is performed on the same $7\times 7$ grid as above. The number of iterations in the DP sweeps is $\ell_{\mathrm{DP}}=5$, with the range $r_{\mathrm{DP}}=5$. The number of nearest-neighbors for the LLR approximation is $N=4$.

The algorithm is implemented in Matlab/Simulink and runs in real time on a remote computer, which communicates via Wi-Fi with the drone. The motion capture system gives very precise position measurements for the drone’s center of gravity in a roughly cubical region. 
The drone comes with a built in stabilization algorithm, which permits the drone to take off/land and hover at a fixed point in space. On top of this algorithm, a cascaded control loop via Wi-Fi has been added as explained next.

\begin{figure}[!htb]
  \centering
  \includegraphics[width=0.3\columnwidth]{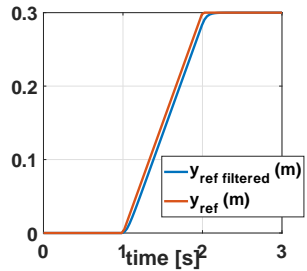}
  \caption{Example reference signal on Y generated based on the high-level action with heading $\pi/2$.}
  \label{fig:ReferenceGen}
\end{figure}

Based on the outer high-level control provided by the learning algorithm, position references are generated in such a way that after $6$\,s the drone is $0.3$\,m away from the current position in the correct direction. Note that the new position will not be exactly the required one, and the learning algorithm runs with the real, updated position at the next step (so it understands that the drone does not move exactly on a grid). The altitude after take off is kept constant at 1\,m. The reference positions on the X and Y axes are generated as ramp, constant-speed signals. Note that the drone points along the X axis, and $p$ contains Y and X positions in this order. First-order reference prefilters with a time constant of $0.1$\,s were also added; see also Figure \ref{fig:ReferenceGen}.

\begin{figure}[!htb]
  \centering
  \includegraphics[width=\columnwidth]{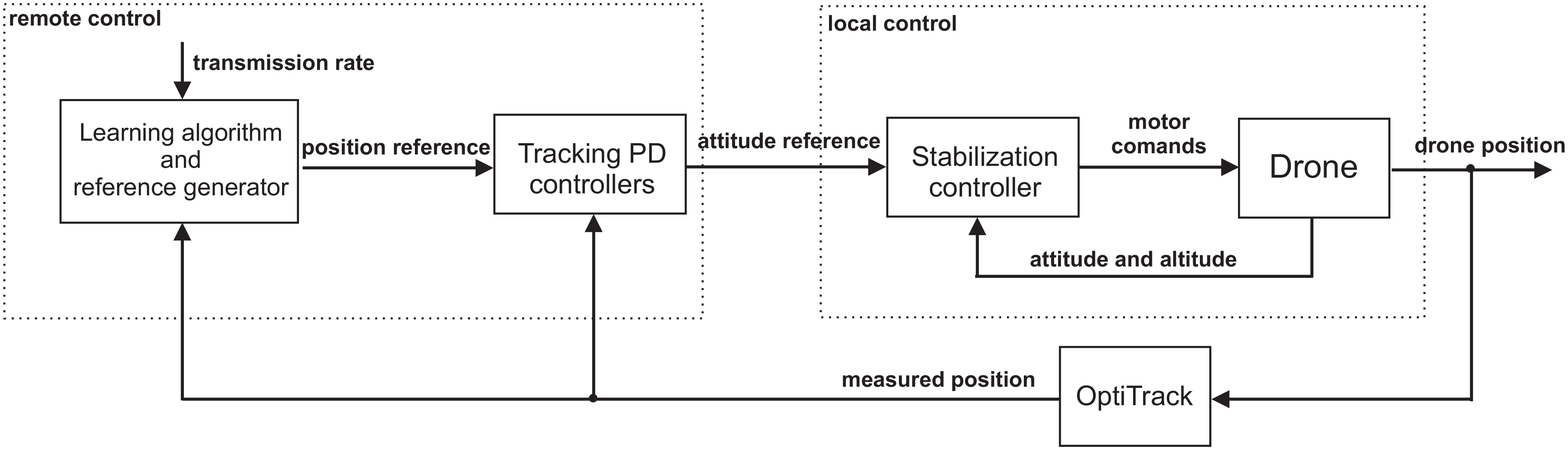}
  \caption{Overview of the control structure}
  \label{fig:ControlStruct}
\end{figure}
These reference positions are then tracked with a mid-level PD control loop, also running on the remote computer; see \figref{fig:ControlStruct}. To design these PD controllers, least-squares identification was performed for the parameters of a second-order model for the dynamics of the drone on each axis, including the low-level feedback loops. The dynamics obtained were $\frac{7.5}{s(1+2s)}$ on the X axis, and $\frac{12.5}{s(1+s)}$ on Y. Then, the PD controllers were designed so as to compensate the time constants of the drone, and with the gains tuned experimentally to reduce the tracking error. The resulting controllers were $0.2 (1+2s)$ and $0.4 (1+s)$ on the two axes, respectively. The sampling period of this mid-level feedback loop if $0.065$\,s. The outputs of this loop are then sent over Wi-Fi as attitude setpoints (pitch and roll) for the inner, low-level control loop, implemented by the producer in the firmware of the drone.

\begin{figure}[!htb]
  \centering
  \includegraphics[width=0.45\textwidth]{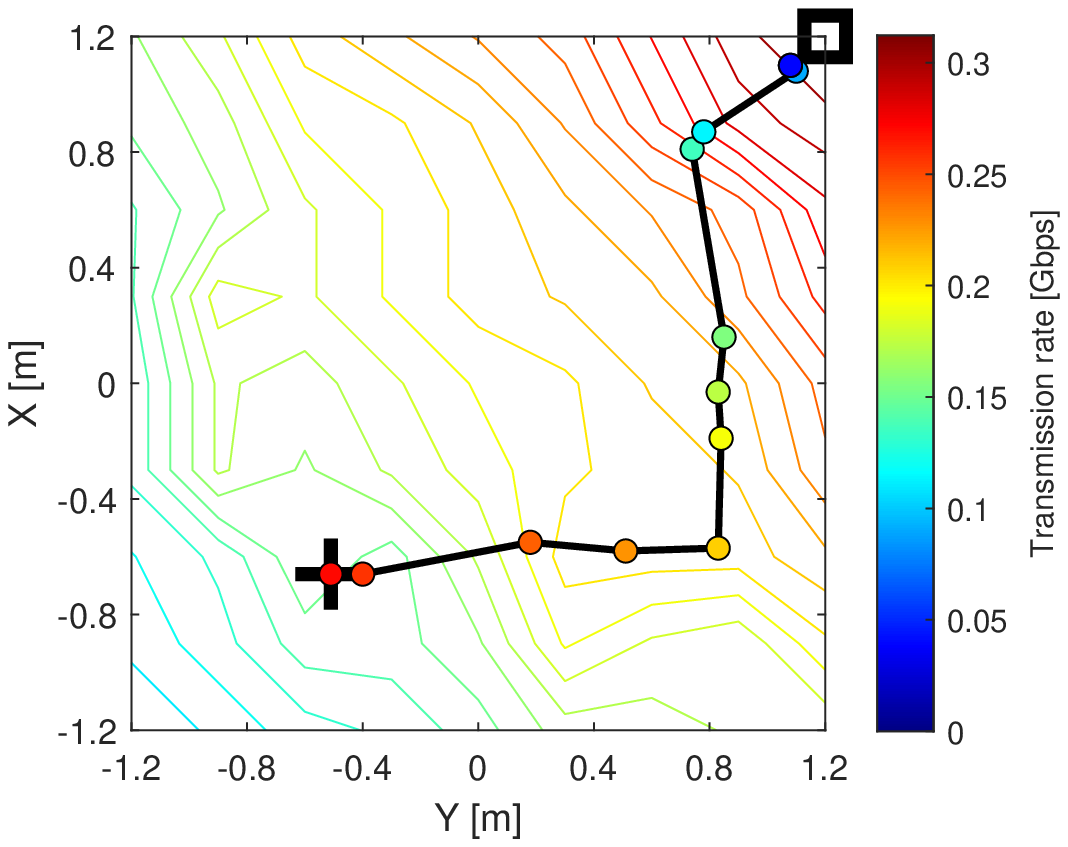}
  \includegraphics[height=0.42\textwidth]{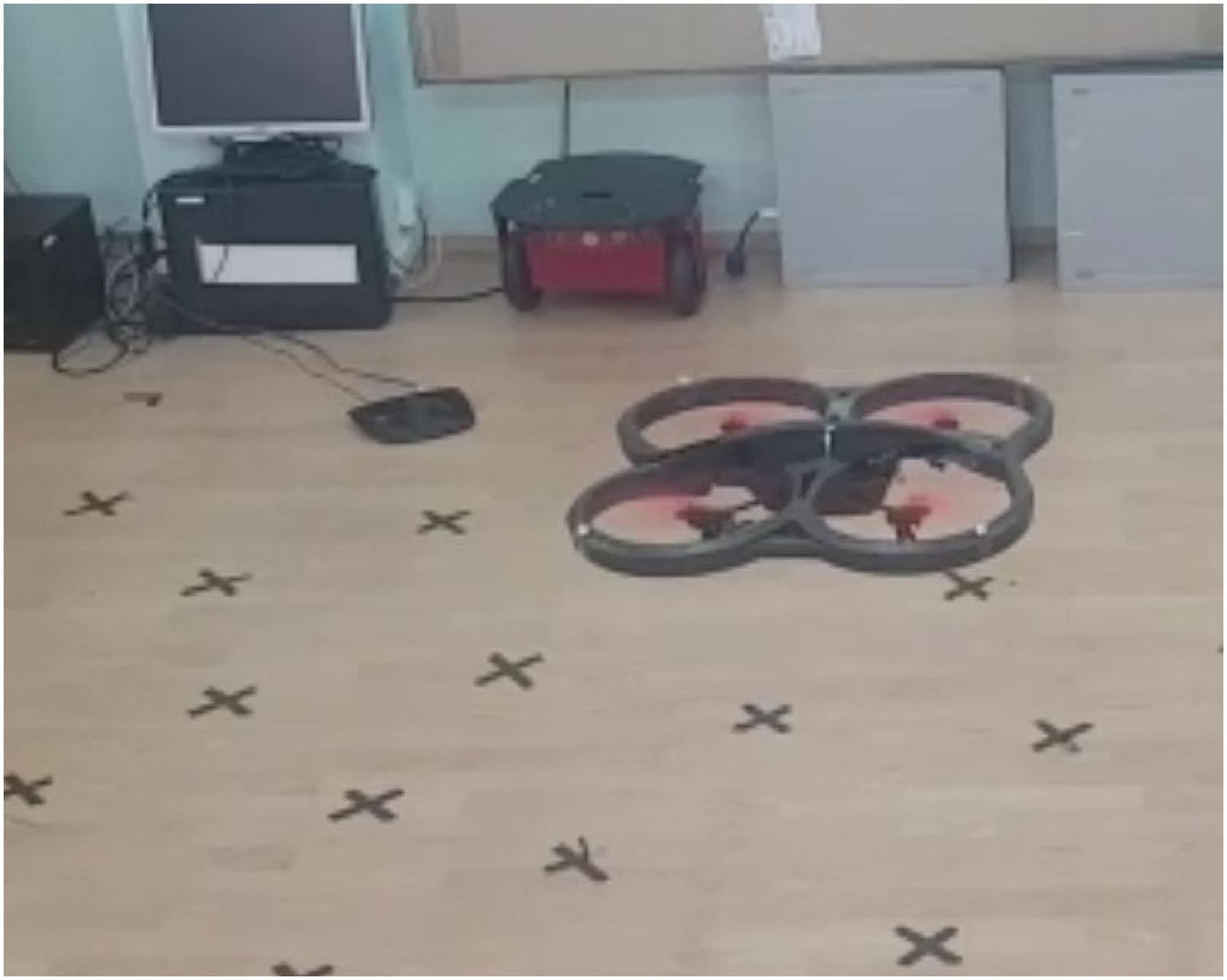}
  \caption{Left: experimental results with a quadcopter drone and the learning algorithm for the transmission problem. The meaning of the contour map, markers etc. is the same as in the simulations above, \rev{and the figure additionally shows (i) the router position with a square and (ii) a color scale for the rate function values. Note that the color has a different meaning for the markers (buffer size) and for the contour plot (transmission rate).} Right: video still of the drone during an experiment.}
  \label{fig:realtimeresult}
\end{figure}

In the particular experimental scenario executed, the transmitter/receiver is a 2.4GHz Wi-Fi router in the upper-right position, $1.2, 1.2$\,m. The level curves represent the transmission rate, in Gbps. Note that the rates are overall larger closer to the router, as expected. The initial buffer value is $b_0= 20$\,Gbit. The drone starts in hovering mode, from the initial position $-0.51,-0.66$\,m. An experimental trajectory is shown in \figref{fig:realtimeresult}, left. This trajectory is qualitatively similar to the one obtained in simulation (see e.g.\ \figref{fig:transmit_traj}, right). In the end, the drone reaches a point very close to the transmitter, with an empty buffer.

The path is not really the ideal concatenation of straight and equal lines on the grid, due to oscillations around the set points. These limitations are unavoidable due to errors introduced by the relatively low-quality drone components and by its inner control loop, coupled with the need to keep the drone in the small volume covered by the positioning system. Nevertheless, the experiment is sufficient to illustrate the overall behavior of the algorithm. \rev{Note that in an outdoor scenario, GPS positioning could be used, and position control errors would likely be negligible compared to the scale of the experiment, in which case these limitations would not apply.}

A video still (from the same experimental scenario, but not the same run as the one in \figref{fig:realtimeresult}, left) is provided in \figref{fig:realtimeresult}, right, and the full video can be accessed online at \url{http://rocon.utcluj.ro/files/commdrone.mp4}.

\section{Conclusions}\label{sec:conclusions}

Two learning-based algorithms were proposed for a mobile robot to transmit data over a wireless network with an unknown rate map: one when the trajectory is free, in which case rectangular obstacles can be handled; and another when the robot must end up at a goal position. Extensive simulations showed that these algorithms achieve good performance, in some cases very close to model-based solutions that require to know the rate function. An illustration with a real UAV was given.

\rev{A relatively simple extension of either algorithm would be to allow acquiring new data while transmitting. Since the algorithms recompute the control law online at each encountered state, they should still work well when the buffer size occasionally increases.} Another interesting extension would be to learn the rate function using Gaussian processes, which in addition to the function estimate also provide an uncertainty at each point, and this uncertainty can be used to drive exploration. It would also help to explicitly take into account random fluctuations in the transmission-problem algorithm; and to handle more general rate function shapes in the navigation-problem algorithm.

\subsection*{Acknowledgments}

This work was supported by a grant of the Romanian Ministry of Research and Innovation, CNCS - UEFISCDI, project number PN-III-P1-1.1-TE-2016-0670, within PNCDI III, and by the Post-Doctoral Programme ``Entrepreneurial competences and excellence research in doctoral and postdoctoral programs - ANTREDOC'', project co-funded from European Social Fund, contract no. 56437/24.07.2019.

Conflict of interest -- none declared.

\section*{References}
\bibliographystyle{elsarticle-harv}
\bibliography{cep}

\begin{thebibliography}{36}
\expandafter\ifx\csname natexlab\endcsname\relax\def\natexlab#1{#1}\fi
\expandafter\ifx\csname url\endcsname\relax
  \def\url#1{\texttt{#1}}\fi
\expandafter\ifx\csname urlprefix\endcsname\relax\def\urlprefix{URL }\fi

\bibitem[{Bertsekas(2012)}]{Bertsekas:12}
Bertsekas, D.~P., 2012. Dynamic Programming and Optimal Control, 4th Edition.
  Vol.~2. Athena Scientific.

\bibitem[{Bristeau et~al.(2011)Bristeau, Callou, Vissi{\`e}re, Petit,
  et~al.}]{Bristeau:11}
Bristeau, P.-J., Callou, F., Vissi{\`e}re, D., Petit, N., et~al., 2011. The
  navigation and control technology inside the ar. drone micro uav. In: 18th
  IFAC World Congress. Vol.~18. pp. 1477--1484.

\bibitem[{Bu\c{s}oniu et~al.(2010)Bu\c{s}oniu, Ernst, {De Schutter}, and
  Babu\v{s}ka}]{aut10}
Bu\c{s}oniu, L., Ernst, D., {De Schutter}, B., Babu\v{s}ka, R., 2010.
  Approximate dynamic programming with a fuzzy parameterization. Automatica
  46~(5), 804--814.

\bibitem[{Bu{\c{s}}oniu et~al.(2019)Bu{\c{s}}oniu, Varma, Mor\u{a}rescu, and
  Lasaulce}]{acc19}
Bu{\c{s}}oniu, L., Varma, V.~S., Mor\u{a}rescu, I.-C., Lasaulce, S., 10--12
  July 2019. Learning-based control for a communicating mobile robot under
  unknown rates. In: Proceedings 2019 Annual American Control Conference
  ({ACC-19}). Philadelphia, USA, pp. 267--272.

\bibitem[{Chatzipanagiotis et~al.(2012)Chatzipanagiotis, Liu, Petropulu, and
  Zavlanos}]{Chatzipanagiotisetal:12}
Chatzipanagiotis, N., Liu, Y., Petropulu, A., Zavlanos, M.~M., 10--13 December
  2012. Controlling groups of mobile beamformers. In: Proceedings 51st IEEE
  Conference on Decision and Control (CDC). Maui, Hawaii, pp. 1984--1989.

\bibitem[{D'Errico(2012)}]{derrico:12}
D'Errico, J., 2012. Bound constrained optimization using fminsearch.
\newline\urlprefix\url{https://www.mathworks.com/matlabcentral/fileexchange/8277-fminsearchbnd-fminsearchcon}

\bibitem[{Fink and Kumar(2010)}]{FinkKumar:10}
Fink, J., Kumar, V., 2010. Online methods for radio signal mapping with mobile
  robots. In: 2010 IEEE International Conference on Robotics and Automation.
  IEEE, pp. 1940--1945.

\bibitem[{Fink et~al.(2012)Fink, Ribeiro, and Kumar}]{Finketal:12}
Fink, J., Ribeiro, A., Kumar, V., 2012. Robust control for mobility and
  wireless communication in cyber--physical systems with application to robot
  teams. Proceedings of the IEEE 100~(1), 164--178.

\bibitem[{Fink et~al.(2013)Fink, Ribeiro, and Kumar}]{Finketal:13}
Fink, J., Ribeiro, A., Kumar, V., 2013. Robust control of mobility and
  communications in autonomous robot teams. IEEE Access 1, 290--309.

\bibitem[{Furtado et~al.(2019)Furtado, Liu, Lai, Lacheray, and
  Desouza-Coelho}]{Furtadoetal:19}
Furtado, J.~S., Liu, H.~H., Lai, G., Lacheray, H., Desouza-Coelho, J., 2019.
  Comparative analysis of optitrack motion capture systems. In: Advances in
  Motion Sensing and Control for Robotic Applications. Springer, pp. 15--31.

\bibitem[{Gangula et~al.(2017)Gangula, de~Kerret, Esrafilian, and
  Gesbert}]{Gangulaetal:17}
Gangula, R., de~Kerret, P., Esrafilian, O., Gesbert, D., Oct 2017. Trajectory
  optimization for mobile access point. In: 51st Asilomar Conference on
  Signals, Systems, and Computers. pp. 1412--1416.

\bibitem[{Ghaffarkhah and Mostofi(2011)}]{GhaffarkhahMostofi:11}
Ghaffarkhah, A., Mostofi, Y., 2011. Communication-aware motion planning in
  mobile networks. IEEE Transactions on Automatic Control 56~(10), 2478--2485.

\bibitem[{Krause et~al.(2008)Krause, Singh, and Guestrin}]{Krauseetal:08}
Krause, A., Singh, A., Guestrin, C., 2008. Near-optimal sensor placements in
  gaussian processes: Theory, efficient algorithms and empirical studies.
  Journal of Machine Learning Research 9~(2), 235--284.

\bibitem[{Licea et~al.(2016)Licea, Varma, Lasaulce, Daafouz, and
  Ghogho}]{Liceaetal:16}
Licea, D.~B., Varma, V.~S., Lasaulce, S., Daafouz, J., Ghogho, M., 26--29
  October 2016. Trajectory planning for energy-efficient vehicles with
  communications constraints. In: Proceedings 2016 International Conference on
  Wireless Networks and Mobile Communications ({WINCOM16}). Fez, Morocco, pp.
  264--270.

\bibitem[{Licea et~al.(2017)Licea, Varma, Lasaulce, Daafouz, Ghogho, and
  McLernon}]{Liceaetal:17}
Licea, D.~B., Varma, V.~S., Lasaulce, S., Daafouz, J., Ghogho, M., McLernon,
  D., 2017. Robust trajectory planning for robotic communications under fading
  channels. In: Ubiquitous Networking: Third International Symposium, UNet
  2017, Casablanca, Morocco, May 9-12, 2017, Revised Selected Papers. Vol.
  10542. Springer, p. 450.

\bibitem[{Lin(1992)}]{Lin:92}
Lin, L.-J., Aug. 1992. Self-improving reactive agents based on reinforcement
  learning, planning and teaching. Machine Learning 8~(3--4), 293--321, special
  issue on reinforcement learning.

\bibitem[{Loh{\'e}ac et~al.(2019)Loh{\'e}ac, S.~Varma, and
  Morarescu}]{Loheacetal:19}
Loh{\'e}ac, J., S.~Varma, V., Morarescu, I.-C., Nov. 2019. {Optimal control for
  a mobile robot with a communication objective}, working paper or preprint.
\newline\urlprefix\url{https://hal.archives-ouvertes.fr/hal-02353582}

\bibitem[{Meera et~al.(2019)Meera, Popovic, Millane, and
  Siegwart}]{Meeraetal:19}
Meera, A.~A., Popovic, M., Millane, A., Siegwart, R., 2019. Obstacle-aware
  adaptive informative path planning for uav-based target search. arXiv
  preprint arXiv:1902.10182.

\bibitem[{Miranda et~al.(2013)Miranda, Abrishambaf, Gomes, Gon{\c{c}}alves,
  Cabral, Tavares, and Monteiro}]{Mirandaetal:13}
Miranda, J., Abrishambaf, R., Gomes, T., Gon{\c{c}}alves, P., Cabral, J.,
  Tavares, A., Monteiro, J., 2013. Path loss exponent analysis in wireless
  sensor networks: Experimental evaluation. In: 2013 11th IEEE International
  Conference on Industrial Informatics (INDIN). IEEE, pp. 54--58.

\bibitem[{Mnih et~al.(2015)Mnih, Kavukcuoglu, Silver, Rusu, Veness, Bellemare,
  Graves, Riedmiller, Fidjeland, Ostrovski, Petersen, Beattie, Sadik,
  Antonoglou, King, Kumaran, Wierstra, Legg, and Hassabis}]{Mnihetal:15}
Mnih, V., Kavukcuoglu, K., Silver, D., Rusu, A.~A., Veness, J., Bellemare,
  M.~G., Graves, A., Riedmiller, M., Fidjeland, A.~K., Ostrovski, G., Petersen,
  S., Beattie, C., Sadik, A., Antonoglou, I., King, H., Kumaran, D., Wierstra,
  D., Legg, S., Hassabis, D., 2015. Human-level control through deep
  reinforcement learning. Nature 518, 529--533.

\bibitem[{Moore and Atkeson(1993)}]{MooreAtkeson:93}
Moore, A.~W., Atkeson, C.~G., 1993. Prioritized sweeping: {R}einforcement
  learning with less data and less time. Machine Learning 13, 103--130.

\bibitem[{Olfati-Saber et~al.(2007)Olfati-Saber, Fax, and
  Murray}]{OlfatiSaberFaxetal:07}
Olfati-Saber, R., Fax, J.~A., Murray, R.~M., 2007. Consensus and cooperation in
  networked multi-agent systems. Proceedings of the {IEEE} 95~(1), 215--233.

\bibitem[{O'Neill et~al.(2017)O'Neill, Delany, and MacNamee}]{ONeilletal:17}
O'Neill, J., Delany, S.~J., MacNamee, B., 2017. Model-free and model-based
  active learning for regression. In: Advances in Computational Intelligence
  Systems. Springer, pp. 375--386.

\bibitem[{Ooi and Schindelhauer(2009)}]{Ooietal:09}
Ooi, C.~C., Schindelhauer, C., 2009. Minimal energy path planning for wireless
  robots. Mobile Networks and Applications 14~(3), 309--321.

\bibitem[{Penumarthi et~al.(2017)Penumarthi, Li, Banfi, Basilico, Amigoni,
  O'Kane, Rekleitis, and Nelakuditi}]{Penumarthietal:17}
Penumarthi, P.~K., Li, A.~Q., Banfi, J., Basilico, N., Amigoni, F., O'Kane, J.,
  Rekleitis, I., Nelakuditi, S., 2017. Multirobot exploration for building
  communication maps with prior from communication models. In: 2017
  International Symposium on Multi-Robot and Multi-Agent Systems (MRS). IEEE,
  pp. 90--96.

\bibitem[{Popovic et~al.(2018)Popovic, Vidal-Calleja, Hitz, Chung, Sa,
  Siegwart, and Nieto}]{Popovicetal:18}
Popovic, M., Vidal-Calleja, T., Hitz, G., Chung, J.~J., Sa, I., Siegwart, R.,
  Nieto, J., 2018. An informative path planning framework for uav-based terrain
  monitoring. arXiv preprint arXiv:1809.03870.

\bibitem[{Rasmussen and Williams(2006)}]{RasmussenWilliams:06}
Rasmussen, C.~E., Williams, C. K.~I., 2006. Gaussian Processes for Machine
  Learning. MIT Press.

\bibitem[{Rizzo et~al.(2019)Rizzo, Lera, and Villarroel}]{Rizzoetal:19}
Rizzo, C., Lera, F., Villarroel, J.~L., 2019. {3-D} fadings structure analysis
  in straight tunnels toward communication, localization, and navigation. IEEE
  Transactions on Antennas and Propagation 67~(9), 6123--6137.

\bibitem[{Rizzo et~al.(2013)Rizzo, Tardioli, Sicignano, Riazuelo, Villarroel,
  and Montano}]{Rizzoetal:13}
Rizzo, C., Tardioli, D., Sicignano, D., Riazuelo, L., Villarroel, J.~L.,
  Montano, L., 2013. Signal-based deployment planning for robot teams in
  tunnel-like fading environments. The International Journal of Robotics
  Research 32~(12), 1381--1397.

\bibitem[{Rooker and Birk(2007)}]{RookerBirk:07}
Rooker, M.~N., Birk, A., 2007. Multi-robot exploration under the constraints of
  wireless networking. Control Engineering Practice 15~(4), 435--445.

\bibitem[{Settles(2009)}]{Settles:09}
Settles, B., 2009. Active learning literature survey. Tech. rep., University of
  Wisconsin-Madison, Department of Computer Sciences.

\bibitem[{Sutton(1990)}]{Sutton:90}
Sutton, R.~S., 21--23 June 1990. Integrated architectures for learning,
  planning, and reacting based on approximating dynamic programming. In:
  Proceedings 7th International Conference on Machine Learning (ICML-90).
  Austin, US, pp. 216--224.

\bibitem[{Sutton and Barto(2018)}]{SuttonBarto:18}
Sutton, R.~S., Barto, A.~G., 2018. Reinforcement Learning: An Introduction, 2nd
  Edition. Adaptive Computation and Machine Learning. A Bradford Book.

\bibitem[{Viseras et~al.(2016)Viseras, Wiedemann, Manss, Magel, Mueller,
  Shutin, and Merino}]{Viserasetal:16}
Viseras, A., Wiedemann, T., Manss, C., Magel, L., Mueller, J., Shutin, D.,
  Merino, L., 2016. Decentralized multi-agent exploration with online-learning
  of gaussian processes. In: 2016 IEEE International Conference on Robotics and
  Automation (ICRA). IEEE, pp. 4222--4229.

\bibitem[{Wu et~al.(2019)Wu, Lin, and Huang}]{Wuetal:19}
Wu, D., Lin, C.-T., Huang, J., 2019. Active learning for regression using
  greedy sampling. Information Sciences 474, 90--105.

\bibitem[{Yan and Mostofi(2013)}]{YanMostofi:12}
Yan, Y., Mostofi, Y., 2013. Co-optimization of communication and motion
  planning of a robotic operation under resource constraints and in fading
  environments. IEEE Transactions on Wireless Communications 12.4 (2013):
  12~(4), 1562--1572.

\end{thebibliography}

\end{document}